\documentclass[sigconf, screen]{acmart}

\usepackage{microtype}
\usepackage{caption}
\usepackage{graphicx}
\usepackage{booktabs} 
\usepackage{graphicx}
\usepackage{amsmath}
\usepackage{hyperref}

\usepackage{multirow}
\usepackage{subcaption}
\usepackage{textcomp}
\usepackage{booktabs}
\usepackage{colortbl}  
\usepackage{xcolor}
\usepackage{array}  
\usepackage{booktabs}

\setcopyright{acmlicensed}
\copyrightyear{2018}
\acmYear{2018}
\acmDOI{XXXXXXX.XXXXXXX}
\acmConference[Conference acronym 'XX]{Make sure to enter the correct
  conference title from your rights confirmation email}{June 03--05,
  2018}{Woodstock, NY}
\acmISBN{978-1-4503-XXXX-X/2018/06}

\begin{document}


\title{Divide-and-Conquer Approach to Holistic Cognition in High-Similarity Contexts with Limited Data}

\author{Shijie Wang$^{1,2}$, Zijian Wang$^2$, Yadan Luo$^2$, 
  Haojie Li$^{1, \ast}$, 
  Zi Huang$^2$, Mahsa Baktashmotlagh$^2$ \\ 
  \textsuperscript{1} Shandong University of Science and Technology, China
  \textsuperscript{2} The University of Queensland, Australia
}

\thanks{$^\ast$Corresponding author.}

\affiliation{
  \institution{}
  \country{}
}
\renewcommand{\shortauthors}{Wang et al.}

\begin{abstract}
  Ultra-fine-grained visual categorization (Ultra-FGVC) aims to classify highly similar subcategories within fine-grained objects using limited training samples. However, holistic yet discriminative cues, such as leaf contours in extremely similar cultivars, remain under-explored in current studies, thereby limiting recognition performance. Though crucial, modeling holistic cues with complex morphological structures typically requires massive training samples, posing significant challenges in data-limited scenarios. 
To address this challenge, we propose a novel Divide-and-Conquer Holistic Cognition Network (DHCNet) that implements a divide-and-conquer strategy by decomposing holistic cues into spatially-associated subtle discrepancies and progressively establishing the holistic cognition process, significantly simplifying holistic cognition while reducing dependency on training data. 
Technically, DHCNet begins by progressively analyzing subtle discrepancies, transitioning from smaller local patches to larger ones using a self-shuffling operation on local regions. Simultaneously, it leverages the unaffected local regions to potentially guide the perception of the original topological structure among the shuffled patches, thereby aiding in the establishment of spatial associations for these discrepancies.
Additionally, DHCNet incorporates the online refinement of these holistic cues discovered from local regions into the training process to iteratively improve their quality. As a result, DHCNet uses these holistic cues as supervisory signals to fine-tune the parameters of the recognition model, thus improving its sensitivity to holistic cues across the entire objects. Extensive evaluations demonstrate that DHCNet achieves remarkable performance on five widely-used Ultra-FGVC datasets.
\end{abstract}



\maketitle

\section{Introduction}

\begin{figure}[!t]
     \centering
     \begin{subfigure}[b]{0.95\linewidth}
         \centering
         \includegraphics[width=\linewidth]{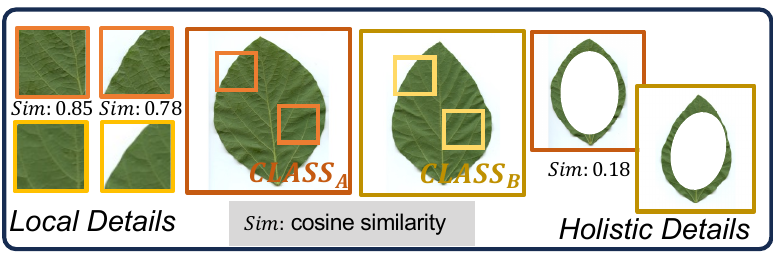}
         \caption{Local details \textit{v.s.} Holistic details}
         \label{fig1a}
     \end{subfigure}
     \hfill
     
     \begin{subfigure}[b]{0.96\linewidth}
         \centering
         \includegraphics[width=\linewidth]{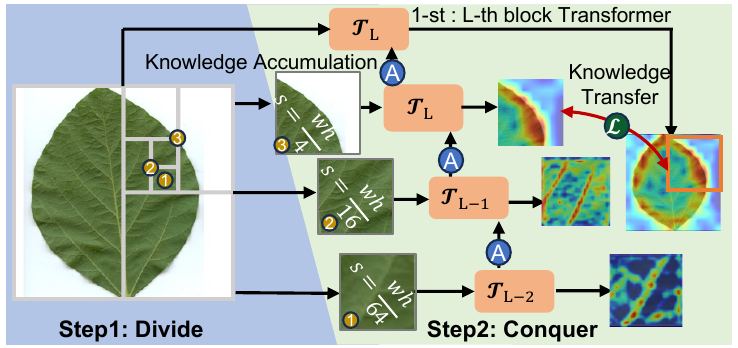}
         \caption{Using divide-to-conquer strategy for inferring holistic cues.}
          \label{fig1b}
     \end{subfigure}
             \caption{
             Motivation of the proposed DHCNet. 
             (a) Holistic cues are significantly more discriminative than local details for identifying extremely similar plant cultivars, as they capture the broader discrepancies and structural patterns across the entire object.
             (b) DHCNet adopts a divide-and-conquer strategy that decomposes holistic perception into hierarchical learning of subtle discriminative details and their spatial configurations.
        }
       
\end{figure}

Ultra-fine-grained visual categorization (Ultra-FGVC) aims at discriminating visually similar subcategories characterized by minute morphological differences. 
This task holds substantial practical value in precision agriculture, including cultivar identification in genomic breeding programs~\cite{qaim2020role, thudi2021genomic} and automated surveillance systems for monitoring cultivar growth~\cite{DBLP:conf/cvpr/YangLZW023,DBLP:journals/tip/PengHZ18, DBLP:journals/pami/JiangWLLZS25}. 
Despite its significance, Ultra-FGVC confronts two fundamental limitations: (1) the existence of extremely subtle visual cues challenging for domain experts to discern, and (2) severe data scarcity where extensive categories contain single-digit annotated samples. 
Such constraints often fail to learn robust and discriminative representations necessary for data-limited Ultra-FGVC.

While contemporary methods ~\cite{DBLP:journals/tcsv/FangJTL24, DBLP:journals/pr/ChenJWDLWW24, DBLP:conf/ijcai/0001WG23, DBLP:conf/wacv/PanYZG23} leverage data augmentation to mitigate data scarcity and employ attention mechanisms to extract local morphological differences, their reliance on local cues proves insufficient against ultra-subtle inter-class variations (Fig.~\ref{fig1a}). Crucially, botanical studies \cite{DBLP:journals/tip/HuJLH12, DBLP:journals/pami/LingJ07, DBLP:journals/tip/WangG14} confirm that holistic morphological cues—such as soybean leaf venation topology and contour geometry—serve as primary discriminative evidence despite occupying minimal pixel areas (Fig.~\ref{fig1a}). 
However, these spatially extensive yet structurally sparse cues present significant learning challenges: modeling their intricate patterns typically requires massive training datasets, rendering conventional deep models ineffective under data-limited conditions ~\cite{DBLP:conf/icml/RadfordKHRGASAM21, DBLP:conf/iclr/YaoHHLNXLLJX22}.
Inspired by human cognitive processes—where global perception arises from the integration of local observations (e.g., \textit{the Chinese parable Blind Men and an Elephant})—we investigate whether holistic cues can be explicitly decomposed into spatially correlated micro-discrepancies. This inquiry motivates a paradigm shift: rather than relying on extensive training data, we advocate for explicitly modeling holistic cognition by breaking down holistic cues into spatially associated subtle differences. This strategy not only alleviates the dependence on large-scale annotated data but also preserves fine-grained discriminative capability.

Motivated by these intuitive analyses, we propose a novel Divide-and-Conquer Holistic Cognition Network (DHCNet), which employs a nested divide-and-conquer strategy to establish holistic cognition of highly similar objects. In the inner-loop divide-and-conquer process, DHCNet incrementally captures subtle discrepancies and their corresponding spatial associations within local regions, enabling the extraction of holistic cues from these local regions.
In the outer-loop divide-and-conquer process, holistic cues derived from local regions serve as supervisory signals to enhance the network's sensitivity to holistic patterns across the entire object. Through this hierarchical decomposition, DHCNet progressively decomposes holistic cues into spatially-associated subtle discrepancies, establishing a comprehensive holistic cognition process, as illustrated in Fig.~\ref{fig1b}.
Additionally, DHCNet incorporates sample augmentation and self-supervisory mechanisms into the divide-and-conquer pipeline, effectively addressing the challenges inherent in data-limited scenarios.

Technically, DHCNet parses the subtle discrepancies present in high-similarity objects and model their spatial associations from limited samples, which are crucial for modeling holistic cues. 
To tackle this, our approach generates augmented images through carefully regulated self-shuffling of local regions, ensuring smooth transitions from fine to broad patches. The network progressively analyzes subtle discrepancies across disruption levels in shuffled regions while leveraging unaffected areas to implicitly guide topological reconstruction within disrupted zones, thus facilitating the establishment of spatial associations for these subtle discrepancies.
More importantly, we design a hierarchical optimization regularization that provides self-supervision signals. This encourages the network to accumulate knowledge of the learned discrepancies and their spatial associations from finer local regions and facilitates the application of these knowledge in broader regions. 
Additionally, DHCNet further integrates online refinement of locally discovered holistic cues into training, iteratively enhancing their quality. These refined cues then serve as supervisory signals to fine-tune recognition parameters, boosting sensitivity to holistic cues across entire ultra-fine-grained objects.

Our contributions are summarized as follows:
\begin{itemize}
    \item 
    To the best of our knowledge, we are the first to employ divide-and-conquer strategy to infer holistic cues with complex morphological structures from single-digit training samples for Ultra-FGVC, providing a new insight in data-efficient ultra-fine-grained recognition.
    \item
    We design DHCNet, which decomposes holistic cues into spatially associated subtle discrepancies, establishes hierarchical cognition through nested divide-and-conquer processing, and transcribes refined holistic representations into recognition model parameters to enhance sensitivity across the entire object.
    \item
    We demonstrate that modeling holistic cues substantially improves the discrimination of highly similar subcategories, with DHCNet achieving a +4.2\% average accuracy gain over state-of-the-art methods~\cite{DBLP:journals/tcsv/FangJTL24} across five large-scale Ultra-FGVC benchmarks.
\end{itemize}

\begin{figure*}[!t]
\begin{center}
   \includegraphics[width=0.95\linewidth]{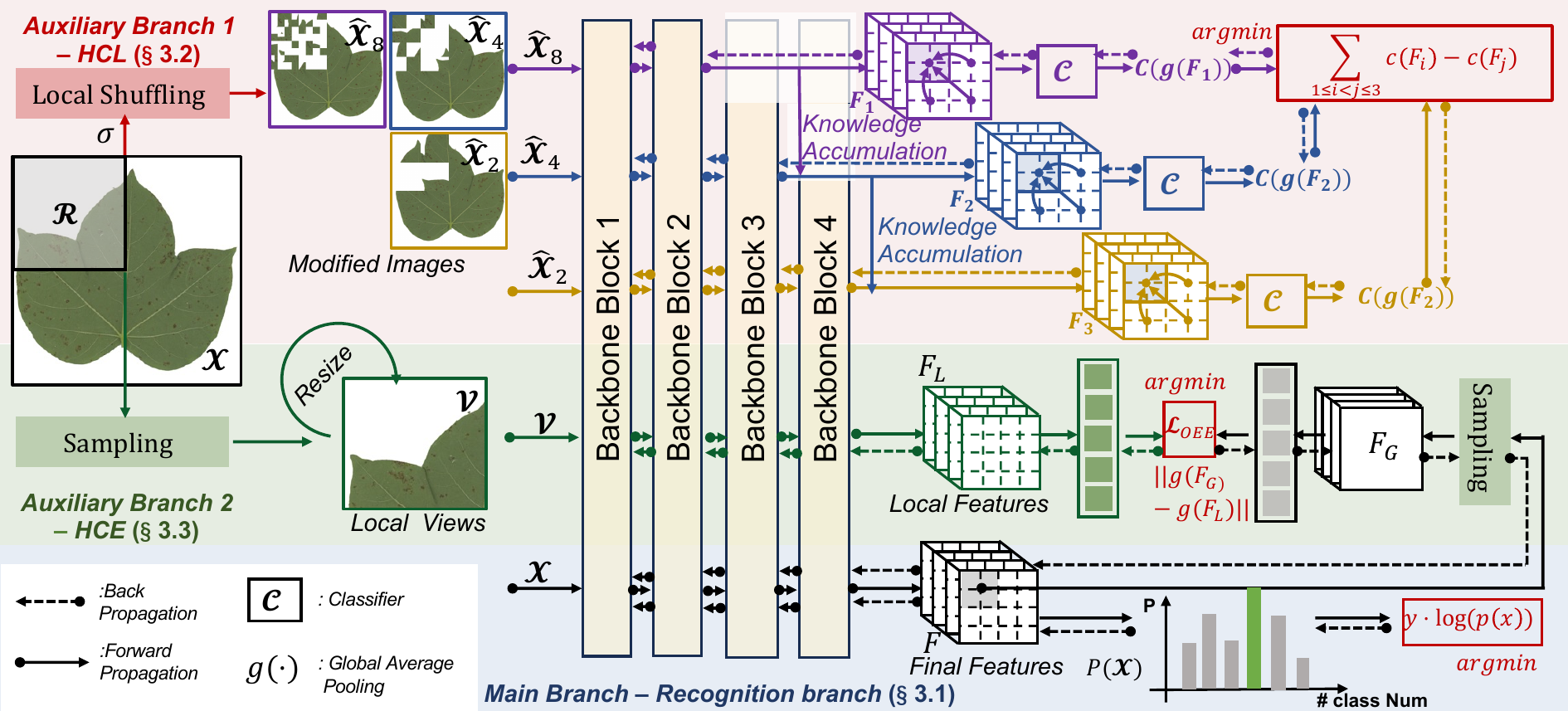}
   \caption{Detailed illustration of {
   \bf Divide-and-Conquer Holistic Cognition}. 
   Our algorithm consists of two auxiliary branches—Holistic Cue Learning (HCL) and Holistic Cognition Expansion (HCE)—along with the main recognition branch. The HCL learns subtle discrepancies and their spatial associations from shuffled regions, while preserving the original topological structure with guidance from unaltered regions. The HCE captures holistic cues from local views to optimize the recognition branch, ensuring that the backbone network incorporates these cues. During testing, only the main recognition branch is used for classification, with HCL and HCE assisting in training to build holistic cognition.
   See \S\textcolor{red}{3} for more details. 
   }
   \label{approach}
\end{center}
\end{figure*}

\section{Related Work}

\textbf{Ultra-fine-grained visual categorization}
\cite{DBLP:conf/aaai/YuZGXY20, DBLP:conf/iccv/YuZ0YX21, DBLP:conf/cvpr/LiuCJQ0P24, DBLP:journals/tcsv/FangJTL24, DBLP:journals/pr/YuZG22}  
identifies high-similarity subcategories within fine-grained objects by employing diverse data augmentation techniques to learn discriminative cues from local regions.
Pioneering work, MORT \cite{DBLP:conf/aaai/YuZGXY20}, 
leverages manually segmented vein structure annotations as supervisory signals to learn object structure representations from local patches.
Recently, there have been some emerging works which aims at a more general conditions without expecting any information about the manual annotations.
UFG-NCD \cite{DBLP:conf/cvpr/LiuCJQ0P24} extracts and utilizes discriminative features from local regions by exploring semantic alignment between regions the along channel direction.
However, previous works predominantly focus on exploring and exploiting discriminative features from local regions by developing diverse data augmentation techniques, often overlooking the valuable information embedded within holistic cues.
To address this limitation, we design the DHCNet network, which employs a divide-and-conquer strategy to establish holistic cognition of highly similar objects and further dig into more discriminative cues.

\noindent\textbf{Divide-to-conquer strategy}
\cite{DBLP:conf/iccv/Wen0CXS23, DBLP:journals/pami/SanakoyeuMTO22, DBLP:conf/cvpr/YangZYWD22} breaks down a complex task into several smaller and simpler sub-tasks, solves each sub-task individually, and then combines their solutions to reach the final solution for the original problem. Utilizing divide-to-conquer strategy makes great progress on various vision tasks, including visual generation \cite{DBLP:conf/cvpr/WangBWFYYZDZWQY24, DBLP:journals/corr/abs-2401-15688, DBLP:journals/corr/abs-2403-06400}, scene understanding \cite{DBLP:journals/ijcv/XiongLLLSC23, DBLP:conf/iccv/NiWYJCH23, DBLP:journals/tip/SongHZCH24}, visual reasoning \cite{DBLP:conf/cvpr/ChenZ23, DBLP:journals/tcsv/WuSXZY23,DBLP:conf/aaai/TianCLYZ24}, and so on.
Most works typically focus on exploring task-related representations from local regions and then generalize them to the entire object, as seen in object counting \cite{DBLP:journals/ijcv/XiongLLLSC23} or shadow removal \cite{DBLP:conf/iccv/GuoWYWW23}.
Inspired by these works using divide-to-conquer strategy, our DHCNet could progressively learn both subtle discrepancies and their spatial associations, starting from smaller local regions and expanding to larger ones, ultimately covering the entire image to infer holistic cues across the entire object.

\textbf{Self-supervised Learning.}
Self-supervised learning~\cite{DBLP:journals/pami/GuiCZCSLT24, DBLP:journals/tkde/LiuJPZZXY23, DBLP:journals/tkde/JiangTL24} has emerged as a transformative paradigm in visual representation learning, leveraging the intrinsic structure of unlabeled data to reduce dependency on large-scale annotated datasets. By designing carefully crafted pretext tasks, this paradigm enables neural networks to learn transferable and discriminative features, thereby enhancing model efficiency and generalization. Recent advances have shown remarkable progress across a wide range of computer vision tasks, including image classification~\cite{DBLP:conf/cvpr/MisraM20, DBLP:conf/cvpr/EricssonGH21}, semantic segmentation~\cite{DBLP:conf/cvpr/WangZKSC20, DBLP:conf/cvpr/ChenYLX22}, 3D representation learning~\cite{DBLP:conf/iccv/HuangXZZ21, DBLP:conf/cvpr/Wang00TPZJ24}, vision-language alignment~\cite{DBLP:conf/cvpr/YangDTXCCZCH22, DBLP:journals/tmm/ZhuangYDQH24}, and domain-specific applications such as autonomous driving~\cite{DBLP:journals/ral/KahnAL21, DBLP:conf/cvpr/SautierPGBBM22} and medical imaging~\cite{DBLP:conf/iccv/AziziMRBFDLKKCN21, DBLP:conf/nips/TalebLDSGBL20}.
Moreover, self-supervised approaches have been successfully applied to model complex intra-object structures, including part-level alignment~\cite{DBLP:journals/tip/PengHZ18, DBLP:conf/eccv/CheFK24, DBLP:conf/cvpr/HeP17} and relational reasoning among object components~\cite{DBLP:conf/cvpr/ZieglerA22, DBLP:conf/nips/PatacchiolaS20}.
Motivated by these advances, we adopt a divide-and-conquer strategy to capture fine-grained details and their spatial dependencies from local regions, which are leveraged as self-supervised signals to propagate local holistic cues toward a global-level understanding.

\section{Methodology}
In this section, we provide a detailed introduction to the Divide-and-Conquer Holistic Cognition Network. 
As shown in Fig.~\ref{approach}, our network is composed of a main branch, \textit{i.e.}, recognition branch, and two auxiliary branches, \textit{i.e.}, holistic cue learning branch, and holistic cognition expansion branch.
The auxiliary branches are designed to assist the recognition branch in learning representations to automatically understand the holistic cues distributed across the entire objects. 
Importantly, the auxiliary branches share the same backbone network as the recognition branch, with no additional learnable parameters introduced. 
At inference, the auxiliary branches are disabled, enhancing computational efficiency by utilizing only the recognition branch.

\subsection{Recognition Branch}
The recognition branch is responsible for extracting basic image features and determining the final object category. Given an input image $\mathbf{X}$ of size $\mathrm{3\times H \times W}$ and its corresponding ground truth one-hot label $\mathbf{y}$, we first obtain the final features $\mathbf{F} = \mathcal{F}_{\mathrm{DNN}}(\mathbf{X})$ from the backbone network. The features are then passed through a classification head to produce the probability vector $\mathbf{p}(\mathbf{X})$, representing the predicted class probabilities.
The loss function of the recognition branch $\mathcal{L}_{\mathrm{CLS}}$ can be written as:
\begin{equation}
    \mathcal{L}_{\mathrm{CLS}} = -\sum_{\mathbf{X} \in \mathcal{X}} \mathrm{y} \cdot \mathrm{log}\mathbf{p}(\mathbf{X}),
\end{equation}
where $\mathcal{X}$ denotes the image set for training.

\subsection{Holistic Cue Learning (HCL)}
Considering the limitation of having few training samples per category, we propose a self-supervised auxiliary branch called Holistic Cue Learning (HCL). 
The core of the HCL is a divide-and-conquer strategy that progressively explores subtle discrepancies and their spatial associations by transitioning from smaller local regions to larger ones using a self-shuffling operation. 
When extracting holistic cues from local regions, HCL boosts sample diversity through self-shuffling operations and incorporates self-supervised signals, thereby mitigating the complexity of modeling holistic cognition in limited-data scenarios.


\textbf{Local shuffling.}
Recent studies show that replacing an input image with a self-shuffled version enhances feature extraction but introduces noise and disrupts object structure, hindering the learning of spatial associations \cite{DBLP:conf/icml/0002FZLZ23}. 
Therefore, we propose a trade-off scheme that shuffles only a local region of the image to explore discriminative discrepancies, while preserving the remaining parts of the image to provide potential guidance for perceiving the original topological structure within the shuffled region.


Formally, given an image $\mathbf{X}$, we select a local region $\mathbf{R}\in\mathbb{R}^{\mathrm{3\times \sqrt{\sigma}H \times \sqrt{\sigma}W}}$ occupying a proportion $\sigma$ of the entire image to be shuffled, enabling the model to learn object discrepancies with spatial associations.
Meanwhile, the remaining $(1-\sigma)$ portion of the image, $\mathbf{\hat{R}}$, remains unaltered to provide potential guidance for perceiving the original topological structure within the shuffled region. 
Note that, to preserve a more complete object region for guidance, we ensure that one vertex of the selected patch aligns with a vertex of the image.

After accquiring $\mathbf{R}$, we equally split into $\mathrm{n} \times \mathrm{n}$ patches which have $\mathrm{3\times \lfloor \frac{\sqrt{\sigma}H }{n} \rfloor\times \lfloor \frac{\sqrt{\sigma}W}{n} } \rfloor$ dimensions.
We adjust the granularity of patches by tuning the hyperparameter $\mathrm{n}$ to guide the network in gradually transitioning from learning smaller-scale discrepancies to larger-range ones.
Specifically, during subsequent training, shuffling operations at different granularities affect various stages of the network. Taking into account the receptive fields at these stages, we set $n  = 2^k$ for $k \in \{1, 2, \cdots, m\} $, to ensure that the shuffled patches \( \mathbf{S} = [\mathbf{S}_1, \mathbf{S}_2, \cdots, \mathbf{S}_m] \) gradually incorporate discrepancies of increasing granularity.
We then integrate these shuffled regions back into the original image to obtain a set of augmented images:
\begin{equation}
  \mathbf{\hat{X}}_k = (\mathbf{S}_k, \mathbf{\hat{R}}), \quad k \in [ 1, 2, \cdots, m].
\end{equation}

\textbf{Extracting holistic cues within local regions.}
Due to the lack of discrepancy annotations, directly feeding all augmented images into large-scale networks like Swin-Transformer can make it challenging to capture discrepancies and their spatial associations, potentially resulting in suboptimal optimization during the learning process.
To address this, we propose a hierarchical optimization strategy that controls the network's layer-wise learning process. 

Initially, the shallow layers are tasked with learning features from a subset of inputs $\mathbf{\hat{X}}_{k_1}$, where ${k_1 \in [\lceil\frac{2m}{3}\rceil + 1, m]}$, to capture discrepancies at a fine-grained level of granularity. Once these local features are effectively learned, the corresponding knowledge—comprising subtle discrepancies and their spatial associations—is accumulated to guide subsequent learning of higher-level holistic cues.
Next, inputs with moderately coarse discrepancies, $\mathbf{\hat{X}}_{k_2}$, where ${k_2 \in [\lceil\frac{m}{3}\rceil + 1, \lceil\frac{2m}{3}\rceil]}$, are introduced to optimize the middle layers for modeling partial structural compositions.
Finally, the remaining inputs, $\mathbf{\hat{X}}_{k_3}$, where ${k_3 \in [1, \lceil\frac{m}{3}\rceil]}$, are used to optimize the deeper layers for capturing more global and abstract structural representations. 
Therefore, the above process can be formulated as:
\begin{equation}
     \mathbf{F}_{\mathrm{i}} = \mathcal{F}_{\mathrm{DNN}}^{\mathrm{(1:(L-i+1))}}(\hat{X}_{k_i}), \quad i \in \left[1, 2, 3\right]
\end{equation}
where $\mathrm{L}$ denotes the stage number of backbone network. The notation $\mathcal{F}_{\mathrm{DNN}}^{\mathrm{(1:(L-i+1))}}$ represents running the backbone network from stage 1 up to stage $\mathrm{(L-i+1)}$, rather than running the entire backbone network, aiming to bypass deeper layers for more targeted parameter optimization.

$\mathcal{F}_{\mathrm{DNN}}$ plays a dual role: it extracts subtle discrepancies from shuffled patches and establishes their spatial associations by utilizing a memorized reference of the unshuffled state to perceive the original topological structure among these patches.
Concretely, the regions chosen for shuffling in each iteration may vary, meaning they do not always undergo the same transformation as in prior iterations. 
Leveraging this variability, the model can memorize the unshuffled states of selected regions from prior iterations. 
By utilizing this memory of the coherent object structure, the model matches correlations among shuffled patches, aiding in the perception of the original topological structure disrupted in the current iteration. 
Consequently, the model is compelled to analyze the visual content within the shuffled regions, fostering an indirect understanding of discrepancies and their spatial associations.


\textbf{Self-supervised signals.}
The divide-and-conquer nature ensures that features derived from higher stages capture both the fine-grained discrepancies and their spatial associations learned in earlier stages, while progressively incorporating broader compositional structures. 
Intuitively, the discriminative power of features  $[\mathbf{F_1}$, $\mathbf{F_2}$, $\cdots$, $\mathbf{F_m}]$ increases, strengthening their influence on defining the classification boundaries. 
To support this, we propose a hierarchical optimization regularization loss, built on top of the cross-entropy loss, to guide the network in gradually optimizing its exploration of holistic cues within local regions:
\begin{equation}
    \mathcal{L}_{\mathrm{HOR}}  = 
     \sum_{1\le \mathrm{i} < \mathrm{j} \le m} \mathrm{log}(\mathcal{C}(\mathbf{F_i})) -  \mathrm{log}(\mathcal{C}(\mathbf{F_j})).
\end{equation}
Here, $\mathcal{C}(\cdot)$ is the confidence function which maps features to the probability corresponding to the ground-truth class.


\subsection{Holistic Cognition Expansion (HCE)}
After the model has learned to recognize holistic cues from local views, we aim to extend this capability to the entire image, enhancing the model's sensitivity to holistic cues distributed across the entire object. 
To accomplish this, we propose an innovative outer-loop divide-and-conquer auxiliary branch. This branch dynamically extracts holistic cues from local views during training, iteratively refines them, and employs these refined cues as supervisory signals to fine-tune the recognition model's parameters. By reassembling holistic cues from local views, this approach enables the model to establish holistic cognition, allowing it to distinguish even highly similar objects accurately.

To ensure consistency with the spatial coverage of holistic cues derived from local regions, we designate each vertex of the input image \(\mathbf{X}\) as a reference point. From these reference points, we extract four local views \(\mathbf{V} = [\mathbf{V}_1, \mathbf{V}_2, \mathbf{V}_3, \mathbf{V}_4]\), each with dimensions \(\mathrm{3 \times \sqrt{\sigma}H \times \sqrt{\sigma}W}\). These four local views collectively cover the entire spatial extent of the input image. 
To further enhance the model's sensitivity to holistic cues, these local views are resized using bilinear interpolation, which improves their alignment with global features.

Then, the local views $\mathbf{V}$ are passed through $\mathcal{F}_{\mathrm{DNN}}$:
\begin{equation}
    \mathbf{F_L} = \mathcal{F}_{\mathrm{DNN}}(\mathbf{V}),
\end{equation}
where $\mathbf{F_L} \in \mathbb{R}^{\mathrm{4\times C\times H \times W}}$ is the holistic cue feature set. 

To effectively incorporate and reassemble these holistic cues, we process the final features $\mathbf{F}$ inferred from the original input $\mathbf{X}$ to align with the holistic cue feature set $\mathbf{F_L}$. 
Concretely, we forward the final features $\mathbf{F}$ into a sampler to extract the corresponding holistic cue feature set $\mathbf{F_G} = [\mathbf{F_G^1}, \mathbf{F_G^2}, \mathbf{F_G^3}, \mathbf{F_G^4}]$ based on the coordinates of the four local views. Specifically, we employ the RoIAlign operation from Mask-RCNN \cite{DBLP:conf/iccv/HeGDG17} to accurately extract the feature set $\mathbf{F_G}$, ensuring precise alignment between the global and local views for optimal integration of holistic cues. 

With the holistic cues including $\mathbf{F_G}$ and $\mathbf{F_L}$, we impose an expansion constraint to extend the model's ability to perceive holistic cues from local views to global ones: 
\begin{equation}
    \mathcal{L}_{\mathrm{EXP}} = \sum_{\mathrm{i}=1}^{4} || \mathrm{g}(\mathbf{F_G^{\mathrm{i}}}) - \mathrm{g}(\mathbf{F_L^{\mathrm{i}}})||,
\end{equation}
where $||\cdot ||$ refers to the Frobenius norm, and $\mathrm{g}(\cdot)$ denotes the global average pooling. 
The expansion constraint could directly penalize the classification model and make its parameters be adjusted through back propagation. Therefore, it ensures that the holistic information inferred from local perspectives is effectively propagated throughout the global feature space, enhancing the model's sensitivity to holistic cues distributed across the entire objects.

\subsection{Overall Training Objective}
The overall training loss for the proposed DHCNet can be formulated as:
\begin{equation}
    \mathcal{L} = \alpha \mathcal{L}_{\mathrm{CLS}} + \beta \mathcal{L}_{\mathrm{HOR}} + \gamma \mathcal{L}_{\mathrm{EXP}} ,
\end{equation}
where $\alpha $, $ \beta $, and $ \gamma$ are the hyper-parameters to weigh the three loss items.

\section{Experiments}

\begin{table}
    \centering
     \setlength{\tabcolsep}{12pt}
    \caption{ Ablation study of the proposed auxiliary branches, \textit{i.e.}, Holistic Cue Learning (HCL) and Holistic Cognition Expansion (HCE) on Cotton80 and SoyLoc datasets, respectively.}
    \begin{tabular}{c c c c  c}
    	\toprule[1pt]
       \multirow{2}{*}{Baseline} & \multirow{2}{*}{HCL}& \multirow{2}{*}{HCE}& \multicolumn{2}{c}{Accuracy} \\
       \cline{4-5}
       &&&Cotton80 & SoyLoc \\
       \toprule[0.7pt]
       \checkmark&&&61.3 & 51.8 \\
       \checkmark& \checkmark &&67.4 & 61.3 \\
        \checkmark&  &\checkmark&65.4 &  60.2\\
       \checkmark&\checkmark&\checkmark&\textbf{70.8} &\textbf{65.8} \\

		\toprule[1pt]
	
    \end{tabular}
    
    \label{tab:ablation_aux}
\end{table}

\begin{table}
    \centering
     \setlength{\tabcolsep}{9pt}
    \caption{ Comparison of performance on Cotton80 (COT) and SoyLoc (SoyL) datasets using different combinations of constraints. We employ the cross-entropy loss $\mathcal{L}_{\mathrm{CE}}$ to replace the proposed constraints $\mathcal{L}_{\mathrm{HOR}}$ and  $\mathcal{L}_{\mathrm{DCE}}$, respectively. }
    \begin{tabular}{c c c c c c c c}
    	\toprule[1pt]
       \multicolumn{2}{c}{HCL}& & \multicolumn{2}{c}{HCE}& & \multicolumn{2}{c}{Accuracy} \\
       \cline{1-2} \cline{ 4-5} \cline{7-8}
       
       $\mathcal{L}_{\mathrm{CE}}$ &  $\mathcal{L}_{\mathrm{HOR}}$ &&$\mathcal{L}_{\mathrm{CE}}$&$\mathcal{L}_{\mathrm{EXP}}$ && COT & SoyL \\
       \toprule[0.7pt]
       \checkmark&&&\checkmark&&&64.3& 56.8\\
       &\checkmark&&\checkmark&&&67.5& 61.6\\
        \checkmark&&&&\checkmark&&65.8& 59.4\\
       &\checkmark&&&\checkmark&&\textbf{70.8} &\textbf{65.8}\\

		\toprule[1pt]
	
    \end{tabular}
    
    \label{tab:ablation_loss}
\end{table}

\begin{table*}
    \centering
     \setlength{\tabcolsep}{14pt}
    \caption{ Quantitative comparison with SOTA methods across five widely-used datasets, i.e., Cotton80, SoyLoc, SoyGene, SoyAgeing, and SoyGlobal. Here, ViT-B \cite{DBLP:conf/iclr/DosovitskiyB0WZ21} denotes the Visual Transformer base model, while Swin-B \cite{DBLP:conf/iccv/LiuL00W0LG21} represents the Swin Transformer base model.}
    \begin{tabular}{l c c c ccc}
    	\toprule[1pt]
       \multirow{2}{*}{Method} & \multirow{2}{*}{Backbone}&  \multicolumn{5}{c}{Top 1 Accuracy (\%)} \\
       \cline{3-7}
       &&Cotton80 & SoyLoc &SoyGene & SoyAgeing & SoyGlobal  \\
       \toprule[0.7pt]
       FDCL-DA $_\mathrm{RP24}$ \cite{DBLP:journals/pr/ChenJWDLWW24} &    ResNet-50 & 43.3&49.8&70.0&76.9&54.2\\
        ADL $_\mathrm{TPAMI21}$ \cite{DBLP:journals/pami/ChoeLS21} &  ResNet-50 & 43.8&34.7&55.2&61.7&39.4\\
       BYOL $_\mathrm{NeurIPS20}$ \cite{DBLP:conf/nips/GrillSATRBDPGAP20}  & ResNet-50 & 52.9 &33.2&60.7&64.8&41.4 \\
       DCL $_\mathrm{CVPR19}$ \cite{DBLP:conf/cvpr/ChenBZM19} &  ResNet-50 & 53.8&45.3&71.4&73.2&42.2\\
       MaskCOV $_\mathrm{PR21}$ \cite{DBLP:journals/pr/YuZGX21} &  ResNet-50 & 58.8&46.2&73.6&75.9&50.3 \\
       CSDNet $_\mathrm{TCSVT24}$ \cite{DBLP:journals/tcsv/FangJTL24}  &  ResNet-50 & 61.7&48.2&66.5&78.0&51.1 \\
       \toprule[0.7pt]
 \rowcolor{gray!30}
       Our DHCNet &  ResNet-50 &  \textbf{64.2}&\textbf{62.8}&\textbf{81.2}&\textbf{82.6}&\textbf{60.1}\\
       \toprule[0.7pt]
       DeiT $_\mathrm{ICML21}$ \cite{DBLP:conf/icml/TouvronCDMSJ21} & ViT-B & 54.2&38.7&66.8&69.5&45.3 \\
       TransFG $_\mathrm{AAAI22}$ \cite{DBLP:conf/aaai/HeCLKYBW22} & ViT-B & 54.6&40.7&22.4&72.2&21.2 \\
       SIM-OFE $_\mathrm{TIP24}$ \cite{DBLP:journals/tip/SunHXP24} & ViT-B & 54.6&25.0&15.5&34.8&70.7 \\
       FDCL-DA $_\mathrm{RP24}$ \cite{DBLP:journals/pr/ChenJWDLWW24} & Swin-B & 62.9&53.3&84.6&85.1&79.7 \\
       CLE-ViT $_\mathrm{IJCAI23}$ \cite{DBLP:conf/ijcai/0001WG23} & Swin-B & 63.3&47.2&78.5&82.1&75.2 \\
       CSDNet $_\mathrm{TCSVT24}$ \cite{DBLP:journals/tcsv/FangJTL24} & Swin-B & 67.9&60.5&86.9&83.2&76.2\\
        \toprule[0.7pt]
 \rowcolor{gray!30}
 Our DHCNet & Swin-B & \textbf{70.8}&\textbf{65.8}&\textbf{88.1}&\textbf{90.0}&\textbf{81.2}\\
		\toprule[1pt]

    \end{tabular}
    
   \label{com}
\end{table*}

\subsection{Experimental Setup}
\textbf{Datasets.} 
We evaluate DHCNet on five ultra-fine-grained datasets \cite{DBLP:conf/aaai/YuZGXY20}, namely, Cotton80, SoyLoc, SoyGene, SoyAgeing, and SoyGlobal.
The Cotton80 dataset comprises 80 cotton species, with a total of 480 images, split into 240 images for training and 240 images for testing.
The SoyLoc dataset encompasses 200 soybean species, totaling 1,200 images, divided into 600 images for training and 600 images for testing.
The SoyAgeing dataset consists of 198 soybean species, with 9,900 images, divided equally into 4,950 images for training and 4,950 images for testing.
The SoyGlobal dataset contains 1,938 soybean species, with 11,628 images, split into 5,814 images for training and 5,814 images for testing.
The SoyGene dataset includes 1,110 soybean species, with 23,906 images, split into 12,763 images for training and 11,143 images for testing.
\textit{These datasets suffer from limited training samples per subcategory, often consisting of only a single-digit number of training images.}

\noindent\textbf{Implementation Details.}
We adopt the Swin Transformer Base \cite{DBLP:conf/iccv/LiuL00W0LG21} as our backbone network, initializing it with pre-trained parameters from ImageNet21k \cite{DBLP:conf/cvpr/DengDSLL009}. The backbone consists of four stages, i.e., the total number of stages $\mathrm{L}$ is set to 4.
The input raw images are resized to $512\times 512$ and cropped into $448 \times 448$. We train our model using SGD optimizer with weight decay of 0.0001, momentum of 0.9, and batch size of 16. We adopt the commonly used data augmentation techniques, \textit{i.e.}, random cropping and erasing, left-right flipping, and color jittering for robust feature representations. Our model is relatively lightweight and is trained end-to-end on one NVIDIA A100 GPU for acceleration. 
The initial learning rate is set to $ 10^{-5} $, with exponential decay of 0.9 after every 5 epochs. The total number of training epochs is set to 200.

\subsection{Ablation Study}
\textbf{Important of auxiliary branches.}
Comprehensive evaluations on the Cotton80 and SoyLoc datasets (Tab.~\ref{tab:ablation_aux}) validate the complementary contributions of the Holistic Cue Learning (HCL) and Holistic Cognition Expansion (HCE) branches. The Swin-Transformer baseline achieves 61.3\% on Cotton80 and 51.8\% on SoyLoc, highlighting its limited capacity for holistic perception despite attending to discriminative regions. 
Introducing the HCL branch significantly improves performance to 67.4\% (+6.1\%) by enabling extensively subtle discrepancy perception and modeling their spatial associations within local regions. 
Adding the HCE branch further enhances accuracy by +4.1\% through global holistic propagation, effectively transferring local structural knowledge across the object. 
Importantly, the synergistic integration of both branches yields peak performance of 70.8\% (+9.5\%) on Cotton80 and 65.8\% (+14.0\%) on SoyLoc, demonstrating that holistic cognition—ranging from fine-grained discrepancy mining to global pattern expansion—is essential for Ultra-FGVC under data-scarce conditions.

\textbf{Optimization Efficacy of Hierarchical Constraints.} Quantitative comparisons in Tab.~\ref{tab:ablation_loss} highlight the critical role of our specialized constraints in facilitating holistic cue modeling under limited data regimes. Replacing both $\mathcal{L}_{\mathrm{HOR}}$ and $\mathcal{L}_{\mathrm{EXP}}$ with standard cross-entropy loss $\mathcal{L}_{\mathrm{CE}}$ (Row 1: 64.3\% on Cotton80 / 56.8\% on SoyLoc) eliminates explicit guidance for hierarchical decomposition, indicating that $\mathcal{L}_{\mathrm{CE}}$ alone is insufficient for capturing micron-scale discrepancies. Introducing $\mathcal{L}_{\mathrm{HOR}}$ (Row 2: 67.5\% / 61.6\%) yields a +3.2\% gain on Cotton80 and +4.8\% on SoyLoc by enforcing self-supervised learning of spatially correlated micro-discrepancies. Further replacing $\mathcal{L}_{\mathrm{CE}}$ with $\mathcal{L}_{\mathrm{EXP}}$ (Row 4: 70.8\% / 65.8\%) leads to an additional improvement of +3.3\% / +4.2\% through object-level propagation of locally mined cues. The combined effect of both constraints results in a +6.5\% (Cotton80) / +9.0pp (SoyLoc) performance boost over the baseline, demonstrating that: (1) $\mathcal{L}_{\mathrm{HOR}}$ effectively guides structural decomposition beyond conventional loss objectives, and (2) $\mathcal{L}_{\mathrm{EXP}}$ is essential for translating local discoveries into globally coherent holistic representations.

\subsection{Comparisons with the State-of-the-Arts}
We conduct comprehensive experiments to evaluate the effectiveness of our proposed DHCNet against state-of-the-art Ultra-FGVC methods. Tab.~\ref{com} presents quantitative comparisons across five widely-used datasets: Cotton80, SoyLoc, SoyGene, SoyAgeing, and SoyGlobal. For fair comparison, we categorize existing approaches into ResNet-based and Transformer-based methods, and evaluate DHCNet using both ResNet-50 and Swin-B backbones.

\textbf{Comparison on ResNet-based Methods.} When employing ResNet-50 as the backbone, our DHCNet achieves remarkable performance improvements across all datasets. Specifically, DHCNet outperforms the strongest ResNet-based competitor CSDNet~\cite{DBLP:journals/tcsv/FangJTL24} by significant margins: +2.5\% on Cotton80 (64.2\% vs. 61.7\%), +14.6\% on SoyLoc (62.8\% vs. 48.2\%), +14.7\% on SoyGene (81.2\% vs. 66.5\%), +4.6\% on SoyAgeing (82.6\% vs. 78.0\%), and +9.0\% on SoyGlobal (60.1\% vs. 51.1\%). These substantial improvements demonstrate the effectiveness of our divide-and-conquer strategy in capturing both local discriminative details and global holistic patterns.

\textbf{Comparison on Transformer-based Methods.} With Swin-B backbone, DHCNet continues to demonstrate superior performance, achieving the best results on all five datasets. Notably, DHCNet surpasses the recent strong competitor CSDNet~\cite{DBLP:journals/tcsv/FangJTL24} by +2.9\% on Cotton80 (70.8\% vs. 67.9\%), +5.3\% on SoyLoc (65.8\% vs. 60.5\%), +1.2\% on SoyGene (88.1\% vs. 86.9\%), +6.8\% on SoyAgeing (90.0\% vs. 83.2\%), and +5.0\% on SoyGlobal (81.2\% vs. 76.2\%). The consistent performance gains across different backbone architectures validate the robustness and effectiveness of our approach.

\textbf{Key Observations.} Several important observations emerge from our experimental results. First, DHCNet achieves superior performance even when using ResNet-50 backbone compared to Transformer-based methods, highlighting the effectiveness of our divide-and-conquer strategy. 
Second, the performance improvements are particularly significant on SoyLoc and SoyAgeing datasets, where holistic morphological cues play crucial roles in distinguishing ultra-fine-grained categories. Third, our approach demonstrates consistent performance gains across different backbone architectures, indicating the generalizability.

\begin{table*}[!t]
    \centering
     \setlength{\tabcolsep}{15pt}
    \caption{ The classification accuracy of the competing models on the five subsets of the SoyAgeing dataset. Here, ViT-B \cite{DBLP:conf/iclr/DosovitskiyB0WZ21} denotes the Visual Transformer base model, while Swin-B \cite{DBLP:conf/iccv/LiuL00W0LG21} represents the Swin Transformer base model.}
    \begin{tabular}{l c c c c ccc}
    	\toprule[1pt]
       \multirow{2}{*}{Method} & \multirow{2}{*}{Backbone}&  \multicolumn{6}{c}{Top 1 Accuracy (\%)} \\
       \cline{3-8}
       &&R1&R3&R4&R5&R6&Average\\
       \toprule[0.7pt]
       SimCLR $_\mathrm{ICML20}$ \cite{DBLP:conf/icml/ChenK0H20} & ResNet-50 & 53.6&45.7&45.4&50.4&35.9&46.2 \\
       BYOL $_\mathrm{NeurIPS20}$ \cite{DBLP:conf/nips/GrillSATRBDPGAP20}  & ResNet-50 & 71.1&66.2&66.2&64.7&56.1&64.8 \\
       DCL $_\mathrm{CVPR19}$ \cite{DBLP:conf/cvpr/ChenBZM19} &  ResNet-50 & 76.9&73.8&76.2&76.2&62.9&73.2\\
       MaskCOV $_\mathrm{PR21}$ \cite{DBLP:journals/pr/YuZGX21} &  ResNet-50 & 79.8&74.7&79.6&78.3&67.0&75.9 \\
       FDCL-DA $_\mathrm{RP24}$ \cite{DBLP:journals/pr/ChenJWDLWW24} &    ResNet-50 & 76.4&76.2&79.1&82.8&70.1&76.9\\
       CSDNet $_\mathrm{TCSVT24}$ \cite{DBLP:journals/tcsv/FangJTL24}  &  ResNet-50 & 81.9&77.5&81.8&79.7&69.2&78.0 \\
       \toprule[0.7pt]
 \rowcolor{gray!30}
       Our DHCNet &  ResNet-50 &  \textbf{84.2}&\textbf{85.6}&\textbf{86.8}&\textbf{84.6}&\textbf{71.6}&\textbf{82.6}\\
       \toprule[0.7pt]
       
       DeiT $_\mathrm{ICML21}$ \cite{DBLP:conf/icml/TouvronCDMSJ21} & ViT-B & 73.0&70.4&69.1&74.7&60.5&69.5 \\
       SIM-OFE $_\mathrm{TIP24}$ \cite{DBLP:journals/tip/SunHXP24} & ViT-B & 69.9&73.2&73.1&73.9&63.2&70.7 \\
       TransFG $_\mathrm{AAAI22}$ \cite{DBLP:conf/aaai/HeCLKYBW22} & ViT-B & 75.0&74.6&74.2&76.2&60.8&72.2 \\
       CLE-ViT $_\mathrm{IJCAI23}$ \cite{DBLP:conf/ijcai/0001WG23} & Swin-B & 80.8&83.3&84.2&86.4&76.0&82.1 \\
       CSDNet $_\mathrm{TCSVT24}$ \cite{DBLP:journals/tcsv/FangJTL24} & Swin-B & 83.8&85.2&85.2&84.9&76.9&83.2\\
       FDCL-DA $_\mathrm{RP24}$ \cite{DBLP:journals/pr/ChenJWDLWW24} & Swin-B & 86.4&86.0&87.7&86.8&78.9&85.1 \\
        \toprule[0.7pt]
 \rowcolor{gray!30}
 Our DHCNet & Swin-B & \textbf{90.6}&\textbf{90.7}&\textbf{91.9}&\textbf{91.9}&\textbf{84.7}&\textbf{90.0}\\

       \toprule[1pt]
       \end{tabular}
       \label{tab:com2}
       \end{table*}

\begin{table}[t]
    \centering
     \setlength{\tabcolsep}{10pt}
    \caption{ Performance on Cotton80 dataset using various shuffled proportions $\sigma$ in HCL.}
    \begin{tabular}{c c c c  c c}
    	\toprule[1pt]
      $\sigma$ & 0\% & 25\% &50\%&75\%& 100\% \\
      \toprule[0.7pt]
      Acc. & 65.4\% &\textbf{70.8\%}&68.1\%&65.3\%&63.3\%\\
		\toprule[1pt]
	
    \end{tabular}
    
    \label{tab:ablation_shuffle}
\end{table}

\begin{table}[t]
    \centering
     \setlength{\tabcolsep}{8pt}
    \caption{ Effect on Cotton80 dataset using different hyper-parameters $m$ in HCL.}
    \begin{tabular}{c c c c  cc c}
    	\toprule[1pt]
      $m$ & 1 & 2 &3&4& 5&6 \\
      \toprule[0.7pt]
      Acc. & 65.4\% & 68.7\% & \textbf{70.8\%} & 69.4\% & 67.3\% & 64.3\% \\
		\toprule[1pt]
    \end{tabular}
    
    \label{tab:m}
\end{table}

\subsection{Comparison Across Soybean Growth Stages}
The SoyAgeing dataset presents a unique challenge in Ultra-FGVC by organizing images into five subsets corresponding to distinct soybean growth stages (R1, R3, R4, R5, and R6), which introduces significant intra-class variations due to morphological and physiological changes during plant development. Table~\ref{tab:com2} presents the comparative results on these five subsets.
Our DHCNet demonstrates remarkable robustness across different growth stages, achieving consistent performance improvements over state-of-the-art methods. With ResNet-50 backbone, DHCNet achieves an average accuracy of 82.6\%, outperforming CSDNet~\cite{DBLP:journals/tcsv/FangJTL24} by +4.6\%. 
When employing Swin-B backbone, DHCNet further improves the average accuracy to 90.0\%, surpassing FDCL-DA~\cite{DBLP:journals/pr/ChenJWDLWW24} by +4.9\%.

Traditional methods such as CSDNet~\cite{DBLP:journals/tcsv/FangJTL24} and FDCL-DA~\cite{DBLP:journals/pr/ChenJWDLWW24}, which primarily focus on local discriminative discrepancies, often struggle to maintain robustness across different cultivation stages. This limitation stems from their reliance on pixel-level cues that can vary significantly across growth stages. In contrast, our DHCNet framework captures discriminative holistic cues intrinsic to soybean cultivars, reducing dependence on such fluctuating visual features.
These results underscore the practical advantages of holistic cognition in Ultra-FGVC scenarios where distinguishing highly similar objects across varying conditions is essential. 

\subsection{Further analyses}

\textbf{Effect of Shuffling Intensity.}
Table~\ref{tab:ablation_shuffle} presents the impact of region shuffling proportion $\sigma$ in HCL on Cotton80 recognition performance. Accuracy exhibits an inverse-U trend with respect to $\sigma$, peaking at 70.8\% when $\sigma = 25\%$, but dropping to 63.3\% under full shuffling ($\sigma = 100\%$). This 7.5\% absolute decline falls below the no-shuffling baseline (65.4\%), indicating that excessive perturbation disrupts feature learning. Notably, performance remains relatively stable within the $\sigma = 25\%\text{–}75\%$ range (variation < 5.5\%), highlighting DHCNet’s robustness to moderate structural perturbations. The optimal result at $\sigma = 25\%$ suggests that moderate shuffling effectively enhances local detail discrimination while preserving essential spatial structures.

\textbf{Effect of self-shuffled degree.} The accuracy in Tab.~\ref{tab:m} exhibits a clear performance peak at $m = 3$ (70.8\%) on the Cotton80 dataset. Smaller $m$ values (e.g., $m = 1, 2$) result in lower accuracy (65.4\%–68.7\%) due to insufficient disruption of subtle features, thereby limiting the model’s ability to capture fine-grained distinctions. In contrast, larger $m$ values (e.g., $m = 4$–$6$) lead to performance degradation (as low as 64.3\%), since excessive fragmentation introduces noisy patterns that obscure meaningful semantic structures. The optimal setting at $m = 3$ achieves a balanced trade-off, enabling fine-grained shuffling to emphasize subtle differences while retaining sufficient contextual information for discriminative learning. These results underscore the importance of granularity control in hierarchical contrastive learning frameworks.

\begin{table}[t]
    \centering
     \setlength{\tabcolsep}{12pt}
    \caption{ Comparison of the recognition performance on Cotton80 dataset with frozen and fine-tuning holistic cues within HCE.}
    \begin{tabular}{l c l}
    	\toprule[1pt]
     Optimization & Parameter & Accuracy (\%) \\
     \toprule[0.7pt]
     Frozen & \textbf{88M}& 69.2\%\\
     \toprule[0.7pt]
     Fine-tuning &  \textbf{88M}& {\bf 70.8\%}$_{+1.6\%}$ \\
	\toprule[1pt]
    \end{tabular}
    
    \label{tab:online}
\end{table}

\begin{table}[t]
    \centering
     \setlength{\tabcolsep}{10pt}
    \caption{ Comparison of the performance on Cotton80 and SoyLoc with Mixup and our divide-and-conquer strategy.}
    \begin{tabular}{l c c}
    	\toprule[1pt]
     Method & Cotton80 & SoyLoc \\
     \toprule[0.7pt]
     Mixup & 60.4 \% & 50.3\% \\
     \toprule[0.7pt]
     Our DHCNet &  {\bf 70.8\%}$_{+10.4\%}$ &  {\bf 65.8\%}$_{+15.5\%}$ \\
	\toprule[1pt]
    \end{tabular}
    
    \label{tab:aug}
\end{table}

\begin{figure}[!t]
\begin{center}
   \includegraphics[width=1.\linewidth]{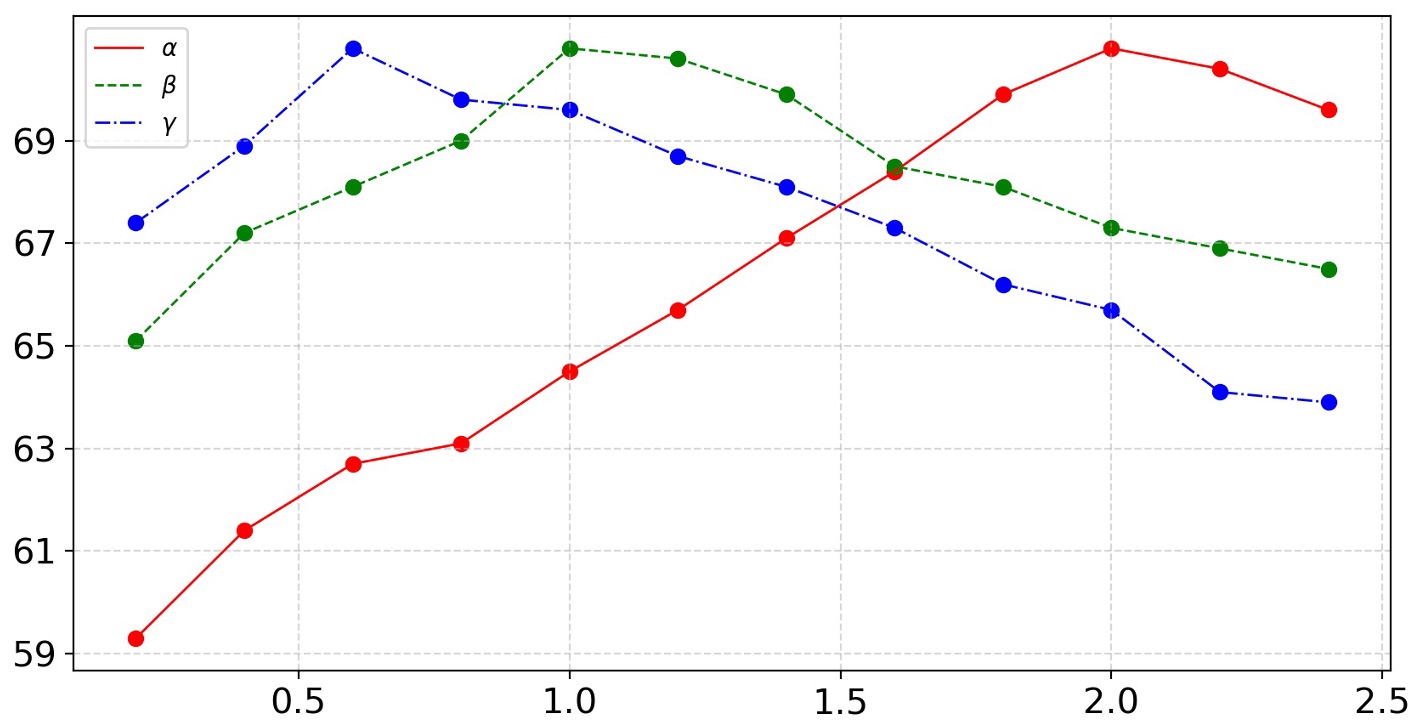}
   \caption{Analyses of hyper-parameters $\alpha$, $\beta$ and $\gamma$ in Eqn. (7). The results denote Top-1 Accuracy on Cotton80.}
   \label{hyper}
\end{center}
\end{figure}

\begin{figure}[!t]
\begin{center}
   \includegraphics[width=1.0\linewidth]{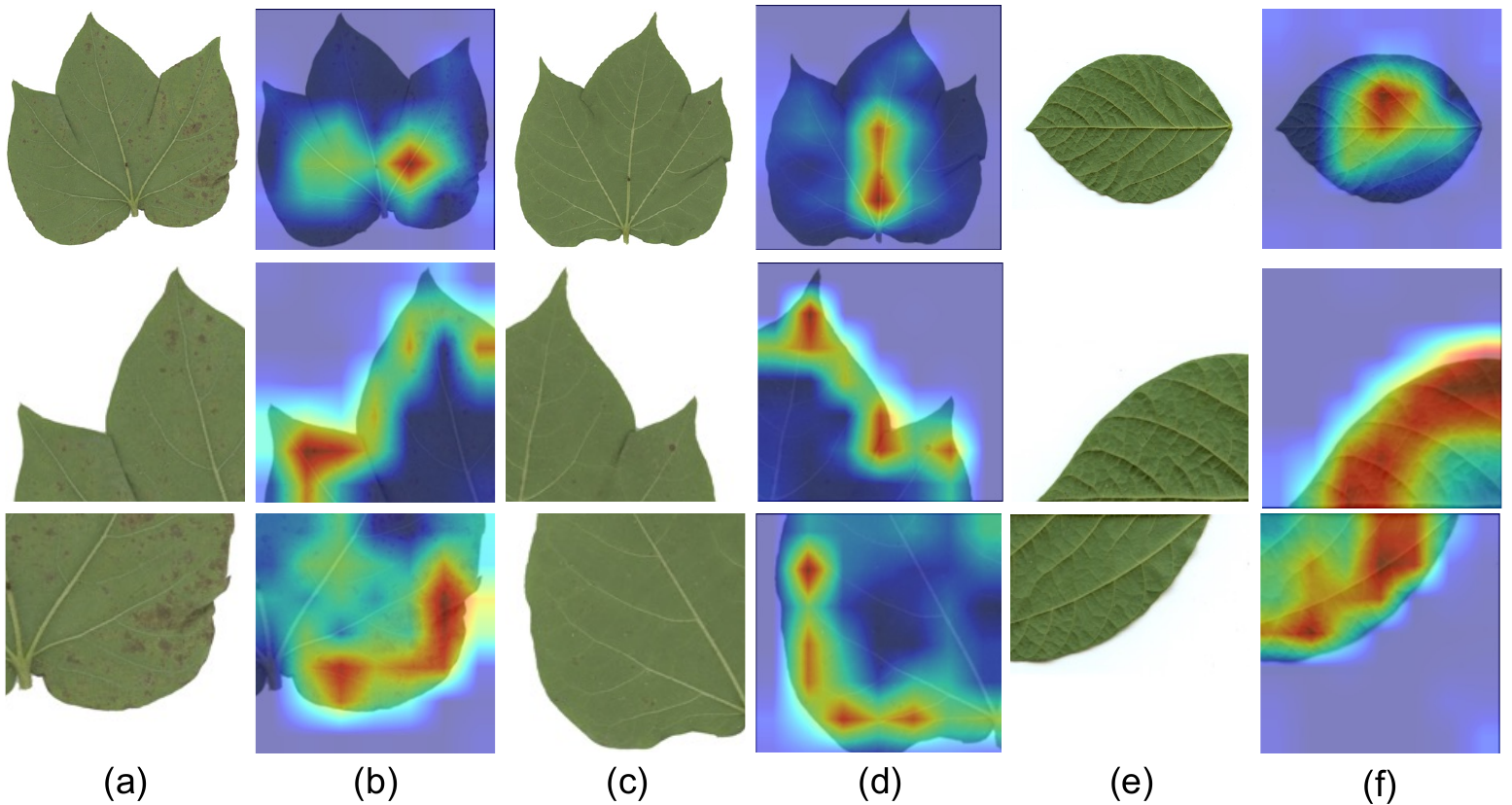}
  
   \caption{
   Visualization of the source of holistic cues. The top row shows the global view. The second and third rows show the multiple image patches created through random cropping. (a)(c)(e) are the input views. (b)(d)(f) show the corresponding response maps.}
    \label{part}
\end{center}
\end{figure}

\textbf{Importance of online refinement of holistic cues.}
Tab.~\ref{tab:online} shows the recognition performance on the Cotton80 dataset with and without online refinement of holistic cues in the HCE branch. In the online setting, we fine-tune the HCE branch, while in the offline setting, the HCE branch is frozen to maintain fixed holistic cues. The results demonstrate that online refinement improves recognition accuracy compared to the offline version. This underscores that optimizing holistic cues in parallel enhances the synergy between locally derived cues and holistic cognition across the entire object.

\textbf{Comparison with Data Augmentation Strategies.}
We conduct comparative experiments to evaluate the effectiveness of our divide-and-conquer strategy against established data augmentation techniques. Table~\ref{tab:aug} presents the performance comparison between Mixup~\cite{DBLP:conf/iclr/ZhangCDL18} and our proposed approach on the Cotton80 and SoyLoc datasets.
The experimental results demonstrate that our DHCNet significantly outperforms the Mixup-based approach across both datasets. 
The Mixup operation blends two images together, which dilutes critical visual cues that are essential for distinguishing highly similar subcategories. In contrast, our divide-and-conquer strategy preserves the structural integrity of individual objects while enhancing discriminative feature learning through controlled region shuffling and spatial association modeling.

\textbf{Hyper-parameter analyses.}
Sensitivity analyses of the hyper-parameters in Eqn. (7) are conducted, and the evaluation results are presented in Fig.~\ref{hyper}. It is observed that the performance of DHCNet is somewhat sensitive to variations in $\alpha$, $\beta$, and $\gamma$. In our experiments, the default values of $\alpha$, $\beta$, and $\gamma$ are set to 2.0, 1.0, and 0.6, respectively, based on empirical tuning for optimal performance.

\begin{figure}[!t]
\begin{center}
   \includegraphics[width=1.0\linewidth]{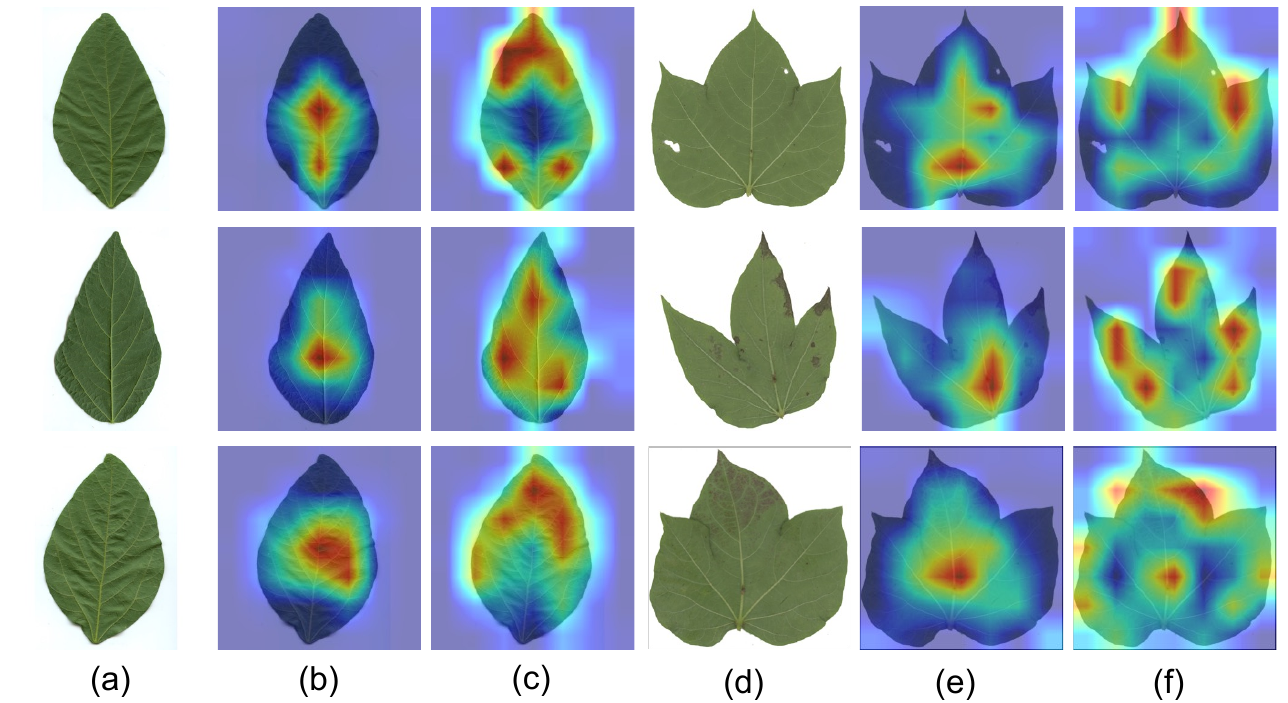}
   \caption{Illustration of class activation maps produced by baseline and our DHCNet. (a)(d) are the input images. (b)(e) are the referred class activation maps provided by baseline. (c)(f) denote class activation maps provided by our DHCNet. }
   \label{whole}
\end{center}
\end{figure}

\textbf{Holistic cue analyses.}
Interpreting the holistic cues learned by DHCNet is challenging, as these cues are optimized within a latent space. To address this, we resort an indirect approach to interpret the holistic cues by visualizing their sources (in Fig.~\ref{part}) to display the content encoded within them. Additionally, we visualize the features influenced by these cues (in Fig.~\ref{whole}) to indirectly trace and understand how these holistic cues contribute to the overall model's decision-making process.
In Fig.~\ref{part}, our DHCNet effectively focuses on spatially continuous information within local views, such as object contours. Based on this observation, we conclude that DHCNet reassembles the holistic cues learned from local views to establish holistic cognition across the entire objects as comprehensively as possible. 
To verify this, we present additional visualization results to demonstrate the influence of holistic cues. The comparison between the baseline and our model is shown in Fig.~\ref{whole}. Unlike the baseline, which tends to focus on fixed center regions, our model captures larger, connected regions. This visualization highlights the significance of holistic cues in this task, and the two auxiliary branches work well on it.

\section{Conclusion}
In this paper, we propose DHCNet, a framework that employs a hierarchical divide-and-conquer strategy to establish holistic cognition for high-similarity objects  in data-limited scenarios.
At the inner-loop divide-and-conquer branch, DHCNet incrementally captures subtle discrepancies and their spatial associations within local regions, enabling the extraction of holistic cues. 
At the outer-loop divide-and-conquer branch, these holistic cues, derived from local regions, serve as supervisory signals to fine-tune the learnable parameters of the recognition model, enhancing the model's sensitivity to holistic cues across the entire object.
Finally, extensive experiments show that our DHCNet outperforms recent state-of-the-art methods by a significant margin on five widely-used Ultra-FGVC benchmarks.

\bibliographystyle{ACM-Reference-Format}
\bibliography{sample-base}


\begin{thebibliography}{64}


\ifx \showCODEN    \undefined \def \showCODEN     #1{\unskip}     \fi
\ifx \showISBNx    \undefined \def \showISBNx     #1{\unskip}     \fi
\ifx \showISBNxiii \undefined \def \showISBNxiii  #1{\unskip}     \fi
\ifx \showISSN     \undefined \def \showISSN      #1{\unskip}     \fi
\ifx \showLCCN     \undefined \def \showLCCN      #1{\unskip}     \fi
\ifx \shownote     \undefined \def \shownote      #1{#1}          \fi
\ifx \showarticletitle \undefined \def \showarticletitle #1{#1}   \fi
\ifx \showURL      \undefined \def \showURL       {\relax}        \fi
\providecommand\bibfield[2]{#2}
\providecommand\bibinfo[2]{#2}
\providecommand\natexlab[1]{#1}
\providecommand\showeprint[2][]{arXiv:#2}

\bibitem[Azizi et~al\mbox{.}(2021)]%
        {DBLP:conf/iccv/AziziMRBFDLKKCN21}
\bibfield{author}{\bibinfo{person}{Shekoofeh Azizi}, \bibinfo{person}{Basil Mustafa}, \bibinfo{person}{Fiona Ryan}, \bibinfo{person}{Zachary Beaver}, \bibinfo{person}{Jan Freyberg}, \bibinfo{person}{Jonathan Deaton}, \bibinfo{person}{Aaron Loh}, \bibinfo{person}{Alan Karthikesalingam}, \bibinfo{person}{Simon Kornblith}, \bibinfo{person}{Ting Chen}, \bibinfo{person}{Vivek Natarajan}, {and} \bibinfo{person}{Mohammad Norouzi}.} \bibinfo{year}{2021}\natexlab{}.
\newblock \showarticletitle{Big Self-Supervised Models Advance Medical Image Classification}. In \bibinfo{booktitle}{\emph{2021 {IEEE/CVF} International Conference on Computer Vision, {ICCV} 2021, Montreal, QC, Canada, October 10-17, 2021}}. \bibinfo{publisher}{{IEEE}}, \bibinfo{pages}{3458--3468}.
\newblock
\href{https://doi.org/10.1109/ICCV48922.2021.00346}{doi:\nolinkurl{10.1109/ICCV48922.2021.00346}}


\bibitem[Che et~al\mbox{.}(2024)]%
        {DBLP:conf/eccv/CheFK24}
\bibfield{author}{\bibinfo{person}{Yuchen Che}, \bibinfo{person}{Ryo Furukawa}, {and} \bibinfo{person}{Asako Kanezaki}.} \bibinfo{year}{2024}\natexlab{}.
\newblock \showarticletitle{OP-Align: Object-Level and Part-Level Alignment for Self-supervised Category-Level Articulated Object Pose Estimation}. In \bibinfo{booktitle}{\emph{Computer Vision - {ECCV} 2024 - 18th European Conference, Milan, Italy, September 29-October 4, 2024, Proceedings, Part {LXXV}}} \emph{(\bibinfo{series}{Lecture Notes in Computer Science}, Vol.~\bibinfo{volume}{15133})}, \bibfield{editor}{\bibinfo{person}{Ales Leonardis}, \bibinfo{person}{Elisa Ricci}, \bibinfo{person}{Stefan Roth}, \bibinfo{person}{Olga Russakovsky}, \bibinfo{person}{Torsten Sattler}, {and} \bibinfo{person}{G{\"{u}}l Varol}} (Eds.). \bibinfo{publisher}{Springer}, \bibinfo{pages}{72--88}.
\newblock
\href{https://doi.org/10.1007/978-3-031-73226-3\_5}{doi:\nolinkurl{10.1007/978-3-031-73226-3\_5}}


\bibitem[Chen et~al\mbox{.}(2024)]%
        {DBLP:journals/pr/ChenJWDLWW24}
\bibfield{author}{\bibinfo{person}{Qiupu Chen}, \bibinfo{person}{Lin Jiao}, \bibinfo{person}{Fenmei Wang}, \bibinfo{person}{Jianming Du}, \bibinfo{person}{Haiyun Liu}, \bibinfo{person}{Xue Wang}, {and} \bibinfo{person}{Rujing Wang}.} \bibinfo{year}{2024}\natexlab{}.
\newblock \showarticletitle{Integrating foreground-background feature distillation and contrastive feature learning for ultra-fine-grained visual classification}.
\newblock \bibinfo{journal}{\emph{Pattern Recognit.}}  \bibinfo{volume}{150} (\bibinfo{year}{2024}), \bibinfo{pages}{110339}.
\newblock
\href{https://doi.org/10.1016/J.PATCOG.2024.110339}{doi:\nolinkurl{10.1016/J.PATCOG.2024.110339}}


\bibitem[Chen et~al\mbox{.}(2022)]%
        {DBLP:conf/cvpr/ChenYLX22}
\bibfield{author}{\bibinfo{person}{Qi Chen}, \bibinfo{person}{Lingxiao Yang}, \bibinfo{person}{Jianhuang Lai}, {and} \bibinfo{person}{Xiaohua Xie}.} \bibinfo{year}{2022}\natexlab{}.
\newblock \showarticletitle{Self-supervised Image-specific Prototype Exploration for Weakly Supervised Semantic Segmentation}. In \bibinfo{booktitle}{\emph{{IEEE/CVF} Conference on Computer Vision and Pattern Recognition, {CVPR} 2022, New Orleans, LA, USA, June 18-24, 2022}}. \bibinfo{publisher}{{IEEE}}, \bibinfo{pages}{4278--4288}.
\newblock
\href{https://doi.org/10.1109/CVPR52688.2022.00425}{doi:\nolinkurl{10.1109/CVPR52688.2022.00425}}


\bibitem[Chen and Zhao(2023)]%
        {DBLP:conf/cvpr/ChenZ23}
\bibfield{author}{\bibinfo{person}{Shi Chen} {and} \bibinfo{person}{Qi Zhao}.} \bibinfo{year}{2023}\natexlab{}.
\newblock \showarticletitle{Divide and Conquer: Answering Questions with Object Factorization and Compositional Reasoning}. In \bibinfo{booktitle}{\emph{{IEEE/CVF} Conference on Computer Vision and Pattern Recognition, {CVPR} 2023, Vancouver, BC, Canada, June 17-24, 2023}}. \bibinfo{publisher}{{IEEE}}, \bibinfo{pages}{6736--6745}.
\newblock
\href{https://doi.org/10.1109/CVPR52729.2023.00651}{doi:\nolinkurl{10.1109/CVPR52729.2023.00651}}


\bibitem[Chen et~al\mbox{.}(2020)]%
        {DBLP:conf/icml/ChenK0H20}
\bibfield{author}{\bibinfo{person}{Ting Chen}, \bibinfo{person}{Simon Kornblith}, \bibinfo{person}{Mohammad Norouzi}, {and} \bibinfo{person}{Geoffrey~E. Hinton}.} \bibinfo{year}{2020}\natexlab{}.
\newblock \showarticletitle{A Simple Framework for Contrastive Learning of Visual Representations}. In \bibinfo{booktitle}{\emph{Proceedings of the 37th International Conference on Machine Learning, {ICML} 2020, 13-18 July 2020, Virtual Event}} \emph{(\bibinfo{series}{Proceedings of Machine Learning Research}, Vol.~\bibinfo{volume}{119})}. \bibinfo{publisher}{{PMLR}}, \bibinfo{pages}{1597--1607}.
\newblock


\bibitem[Chen et~al\mbox{.}(2019)]%
        {DBLP:conf/cvpr/ChenBZM19}
\bibfield{author}{\bibinfo{person}{Yue Chen}, \bibinfo{person}{Yalong Bai}, \bibinfo{person}{Wei Zhang}, {and} \bibinfo{person}{Tao Mei}.} \bibinfo{year}{2019}\natexlab{}.
\newblock \showarticletitle{Destruction and Construction Learning for Fine-Grained Image Recognition}. In \bibinfo{booktitle}{\emph{{IEEE} Conference on Computer Vision and Pattern Recognition, {CVPR} 2019, Long Beach, CA, USA, June 16-20, 2019}}. \bibinfo{publisher}{Computer Vision Foundation / {IEEE}}, \bibinfo{pages}{5157--5166}.
\newblock
\href{https://doi.org/10.1109/CVPR.2019.00530}{doi:\nolinkurl{10.1109/CVPR.2019.00530}}


\bibitem[Choe et~al\mbox{.}(2021)]%
        {DBLP:journals/pami/ChoeLS21}
\bibfield{author}{\bibinfo{person}{Junsuk Choe}, \bibinfo{person}{Seungho Lee}, {and} \bibinfo{person}{Hyunjung Shim}.} \bibinfo{year}{2021}\natexlab{}.
\newblock \showarticletitle{Attention-Based Dropout Layer for Weakly Supervised Single Object Localization and Semantic Segmentation}.
\newblock \bibinfo{journal}{\emph{{IEEE} Trans. Pattern Anal. Mach. Intell.}} \bibinfo{volume}{43}, \bibinfo{number}{12} (\bibinfo{year}{2021}), \bibinfo{pages}{4256--4271}.
\newblock
\href{https://doi.org/10.1109/TPAMI.2020.2999099}{doi:\nolinkurl{10.1109/TPAMI.2020.2999099}}


\bibitem[Deng et~al\mbox{.}(2009)]%
        {DBLP:conf/cvpr/DengDSLL009}
\bibfield{author}{\bibinfo{person}{Jia Deng}, \bibinfo{person}{Wei Dong}, \bibinfo{person}{Richard Socher}, \bibinfo{person}{Li{-}Jia Li}, \bibinfo{person}{Kai Li}, {and} \bibinfo{person}{Li Fei{-}Fei}.} \bibinfo{year}{2009}\natexlab{}.
\newblock \showarticletitle{ImageNet: {A} large-scale hierarchical image database}. In \bibinfo{booktitle}{\emph{2009 {IEEE} Computer Society Conference on Computer Vision and Pattern Recognition {(CVPR} 2009), 20-25 June 2009, Miami, Florida, {USA}}}. \bibinfo{publisher}{{IEEE} Computer Society}, \bibinfo{pages}{248--255}.
\newblock
\href{https://doi.org/10.1109/CVPR.2009.5206848}{doi:\nolinkurl{10.1109/CVPR.2009.5206848}}


\bibitem[Dosovitskiy et~al\mbox{.}(2021)]%
        {DBLP:conf/iclr/DosovitskiyB0WZ21}
\bibfield{author}{\bibinfo{person}{Alexey Dosovitskiy}, \bibinfo{person}{Lucas Beyer}, \bibinfo{person}{Alexander Kolesnikov}, \bibinfo{person}{Dirk Weissenborn}, \bibinfo{person}{Xiaohua Zhai}, \bibinfo{person}{Thomas Unterthiner}, \bibinfo{person}{Mostafa Dehghani}, \bibinfo{person}{Matthias Minderer}, \bibinfo{person}{Georg Heigold}, \bibinfo{person}{Sylvain Gelly}, \bibinfo{person}{Jakob Uszkoreit}, {and} \bibinfo{person}{Neil Houlsby}.} \bibinfo{year}{2021}\natexlab{}.
\newblock \showarticletitle{An Image is Worth 16x16 Words: Transformers for Image Recognition at Scale}. In \bibinfo{booktitle}{\emph{9th International Conference on Learning Representations, {ICLR} 2021, Virtual Event, Austria, May 3-7, 2021}}. \bibinfo{publisher}{OpenReview.net}.
\newblock


\bibitem[Ericsson et~al\mbox{.}(2021)]%
        {DBLP:conf/cvpr/EricssonGH21}
\bibfield{author}{\bibinfo{person}{Linus Ericsson}, \bibinfo{person}{Henry Gouk}, {and} \bibinfo{person}{Timothy~M. Hospedales}.} \bibinfo{year}{2021}\natexlab{}.
\newblock \showarticletitle{How Well Do Self-Supervised Models Transfer?}. In \bibinfo{booktitle}{\emph{{IEEE} Conference on Computer Vision and Pattern Recognition, {CVPR} 2021, virtual, June 19-25, 2021}}. \bibinfo{publisher}{Computer Vision Foundation / {IEEE}}, \bibinfo{pages}{5414--5423}.
\newblock
\href{https://doi.org/10.1109/CVPR46437.2021.00537}{doi:\nolinkurl{10.1109/CVPR46437.2021.00537}}


\bibitem[Fang et~al\mbox{.}(2024)]%
        {DBLP:journals/tcsv/FangJTL24}
\bibfield{author}{\bibinfo{person}{Ziye Fang}, \bibinfo{person}{Xin Jiang}, \bibinfo{person}{Hao Tang}, {and} \bibinfo{person}{Zechao Li}.} \bibinfo{year}{2024}\natexlab{}.
\newblock \showarticletitle{Learning Contrastive Self-Distillation for Ultra-Fine-Grained Visual Categorization Targeting Limited Samples}.
\newblock \bibinfo{journal}{\emph{{IEEE} Trans. Circuits Syst. Video Technol.}} \bibinfo{volume}{34}, \bibinfo{number}{8} (\bibinfo{year}{2024}), \bibinfo{pages}{7135--7148}.
\newblock
\href{https://doi.org/10.1109/TCSVT.2024.3370731}{doi:\nolinkurl{10.1109/TCSVT.2024.3370731}}


\bibitem[Grill et~al\mbox{.}(2020)]%
        {DBLP:conf/nips/GrillSATRBDPGAP20}
\bibfield{author}{\bibinfo{person}{Jean{-}Bastien Grill}, \bibinfo{person}{Florian Strub}, \bibinfo{person}{Florent Altch{\'{e}}}, \bibinfo{person}{Corentin Tallec}, \bibinfo{person}{Pierre~H. Richemond}, \bibinfo{person}{Elena Buchatskaya}, \bibinfo{person}{Carl Doersch}, \bibinfo{person}{Bernardo~{\'{A}}vila Pires}, \bibinfo{person}{Zhaohan Guo}, \bibinfo{person}{Mohammad~Gheshlaghi Azar}, \bibinfo{person}{Bilal Piot}, \bibinfo{person}{Koray Kavukcuoglu}, \bibinfo{person}{R{\'{e}}mi Munos}, {and} \bibinfo{person}{Michal Valko}.} \bibinfo{year}{2020}\natexlab{}.
\newblock \showarticletitle{Bootstrap Your Own Latent - {A} New Approach to Self-Supervised Learning}. In \bibinfo{booktitle}{\emph{Advances in Neural Information Processing Systems 33: Annual Conference on Neural Information Processing Systems 2020, NeurIPS 2020, December 6-12, 2020, virtual}}, \bibfield{editor}{\bibinfo{person}{Hugo Larochelle}, \bibinfo{person}{Marc'Aurelio Ranzato}, \bibinfo{person}{Raia Hadsell}, \bibinfo{person}{Maria{-}Florina Balcan}, {and} \bibinfo{person}{Hsuan{-}Tien Lin}} (Eds.).
\newblock


\bibitem[Gui et~al\mbox{.}(2024)]%
        {DBLP:journals/pami/GuiCZCSLT24}
\bibfield{author}{\bibinfo{person}{Jie Gui}, \bibinfo{person}{Tuo Chen}, \bibinfo{person}{Jing Zhang}, \bibinfo{person}{Qiong Cao}, \bibinfo{person}{Zhenan Sun}, \bibinfo{person}{Hao Luo}, {and} \bibinfo{person}{Dacheng Tao}.} \bibinfo{year}{2024}\natexlab{}.
\newblock \showarticletitle{A Survey on Self-Supervised Learning: Algorithms, Applications, and Future Trends}.
\newblock \bibinfo{journal}{\emph{{IEEE} Trans. Pattern Anal. Mach. Intell.}} \bibinfo{volume}{46}, \bibinfo{number}{12} (\bibinfo{year}{2024}), \bibinfo{pages}{9052--9071}.
\newblock
\href{https://doi.org/10.1109/TPAMI.2024.3415112}{doi:\nolinkurl{10.1109/TPAMI.2024.3415112}}


\bibitem[Guo et~al\mbox{.}(2023)]%
        {DBLP:conf/iccv/GuoWYWW23}
\bibfield{author}{\bibinfo{person}{Lanqing Guo}, \bibinfo{person}{Chong Wang}, \bibinfo{person}{Wenhan Yang}, \bibinfo{person}{Yufei Wang}, {and} \bibinfo{person}{Bihan Wen}.} \bibinfo{year}{2023}\natexlab{}.
\newblock \showarticletitle{Boundary-Aware Divide and Conquer: {A} Diffusion-based Solution for Unsupervised Shadow Removal}. In \bibinfo{booktitle}{\emph{{IEEE/CVF} International Conference on Computer Vision, {ICCV} 2023, Paris, France, October 1-6, 2023}}. \bibinfo{publisher}{{IEEE}}, \bibinfo{pages}{12999--13008}.
\newblock
\href{https://doi.org/10.1109/ICCV51070.2023.01199}{doi:\nolinkurl{10.1109/ICCV51070.2023.01199}}


\bibitem[He et~al\mbox{.}(2022)]%
        {DBLP:conf/aaai/HeCLKYBW22}
\bibfield{author}{\bibinfo{person}{Ju He}, \bibinfo{person}{Jieneng Chen}, \bibinfo{person}{Shuai Liu}, \bibinfo{person}{Adam Kortylewski}, \bibinfo{person}{Cheng Yang}, \bibinfo{person}{Yutong Bai}, {and} \bibinfo{person}{Changhu Wang}.} \bibinfo{year}{2022}\natexlab{}.
\newblock \showarticletitle{TransFG: {A} Transformer Architecture for Fine-Grained Recognition}. In \bibinfo{booktitle}{\emph{Thirty-Sixth {AAAI} Conference on Artificial Intelligence, {AAAI} 2022, Thirty-Fourth Conference on Innovative Applications of Artificial Intelligence, {IAAI} 2022, The Twelveth Symposium on Educational Advances in Artificial Intelligence, {EAAI} 2022 Virtual Event, February 22 - March 1, 2022}}. \bibinfo{publisher}{{AAAI} Press}, \bibinfo{pages}{852--860}.
\newblock
\href{https://doi.org/10.1609/AAAI.V36I1.19967}{doi:\nolinkurl{10.1609/AAAI.V36I1.19967}}


\bibitem[He et~al\mbox{.}(2017)]%
        {DBLP:conf/iccv/HeGDG17}
\bibfield{author}{\bibinfo{person}{Kaiming He}, \bibinfo{person}{Georgia Gkioxari}, \bibinfo{person}{Piotr Doll{\'{a}}r}, {and} \bibinfo{person}{Ross~B. Girshick}.} \bibinfo{year}{2017}\natexlab{}.
\newblock \showarticletitle{Mask {R-CNN}}. In \bibinfo{booktitle}{\emph{{IEEE} International Conference on Computer Vision, {ICCV} 2017, Venice, Italy, October 22-29, 2017}}. \bibinfo{publisher}{{IEEE} Computer Society}, \bibinfo{pages}{2980--2988}.
\newblock
\href{https://doi.org/10.1109/ICCV.2017.322}{doi:\nolinkurl{10.1109/ICCV.2017.322}}


\bibitem[He and Peng(2017)]%
        {DBLP:conf/cvpr/HeP17}
\bibfield{author}{\bibinfo{person}{Xiangteng He} {and} \bibinfo{person}{Yuxin Peng}.} \bibinfo{year}{2017}\natexlab{}.
\newblock \showarticletitle{Fine-Grained Image Classification via Combining Vision and Language}. In \bibinfo{booktitle}{\emph{2017 {IEEE} Conference on Computer Vision and Pattern Recognition, {CVPR} 2017, Honolulu, HI, USA, July 21-26, 2017}}. \bibinfo{publisher}{{IEEE} Computer Society}, \bibinfo{pages}{7332--7340}.
\newblock
\href{https://doi.org/10.1109/CVPR.2017.775}{doi:\nolinkurl{10.1109/CVPR.2017.775}}


\bibitem[Hu et~al\mbox{.}(2012)]%
        {DBLP:journals/tip/HuJLH12}
\bibfield{author}{\bibinfo{person}{Rong{-}Xiang Hu}, \bibinfo{person}{Wei Jia}, \bibinfo{person}{Haibin Ling}, {and} \bibinfo{person}{Deshuang Huang}.} \bibinfo{year}{2012}\natexlab{}.
\newblock \showarticletitle{Multiscale Distance Matrix for Fast Plant Leaf Recognition}.
\newblock \bibinfo{journal}{\emph{{IEEE} Trans. Image Process.}} \bibinfo{volume}{21}, \bibinfo{number}{11} (\bibinfo{year}{2012}), \bibinfo{pages}{4667--4672}.
\newblock


\bibitem[Huang et~al\mbox{.}(2021)]%
        {DBLP:conf/iccv/HuangXZZ21}
\bibfield{author}{\bibinfo{person}{Siyuan Huang}, \bibinfo{person}{Yichen Xie}, \bibinfo{person}{Song{-}Chun Zhu}, {and} \bibinfo{person}{Yixin Zhu}.} \bibinfo{year}{2021}\natexlab{}.
\newblock \showarticletitle{Spatio-temporal Self-Supervised Representation Learning for 3D Point Clouds}. In \bibinfo{booktitle}{\emph{2021 {IEEE/CVF} International Conference on Computer Vision, {ICCV} 2021, Montreal, QC, Canada, October 10-17, 2021}}. \bibinfo{publisher}{{IEEE}}, \bibinfo{pages}{6515--6525}.
\newblock
\href{https://doi.org/10.1109/ICCV48922.2021.00647}{doi:\nolinkurl{10.1109/ICCV48922.2021.00647}}


\bibitem[Jia and Tan(2024)]%
        {DBLP:journals/corr/abs-2403-06400}
\bibfield{author}{\bibinfo{person}{Yuhao Jia} {and} \bibinfo{person}{Wenhan Tan}.} \bibinfo{year}{2024}\natexlab{}.
\newblock \showarticletitle{DivCon: Divide and Conquer for Progressive Text-to-Image Generation}.
\newblock \bibinfo{journal}{\emph{CoRR}}  \bibinfo{volume}{abs/2403.06400} (\bibinfo{year}{2024}).
\newblock
\showeprint[arXiv]{2403.06400}
\href{https://doi.org/10.48550/ARXIV.2403.06400}{doi:\nolinkurl{10.48550/ARXIV.2403.06400}}


\bibitem[Jiang et~al\mbox{.}(2025)]%
        {DBLP:journals/pami/JiangWLLZS25}
\bibfield{author}{\bibinfo{person}{Chenyi Jiang}, \bibinfo{person}{Shidong Wang}, \bibinfo{person}{Yang Long}, \bibinfo{person}{Zechao Li}, \bibinfo{person}{Haofeng Zhang}, {and} \bibinfo{person}{Ling Shao}.} \bibinfo{year}{2025}\natexlab{}.
\newblock \showarticletitle{Imaginary-Connected Embedding in Complex Space for Unseen Attribute-Object Discrimination}.
\newblock \bibinfo{journal}{\emph{{IEEE} Trans. Pattern Anal. Mach. Intell.}} \bibinfo{volume}{47}, \bibinfo{number}{3} (\bibinfo{year}{2025}), \bibinfo{pages}{1395--1413}.
\newblock


\bibitem[Jiang et~al\mbox{.}(2024)]%
        {DBLP:journals/tkde/JiangTL24}
\bibfield{author}{\bibinfo{person}{Xin Jiang}, \bibinfo{person}{Hao Tang}, {and} \bibinfo{person}{Zechao Li}.} \bibinfo{year}{2024}\natexlab{}.
\newblock \showarticletitle{Global Meets Local: Dual Activation Hashing Network for Large-Scale Fine-Grained Image Retrieval}.
\newblock \bibinfo{journal}{\emph{{IEEE} Trans. Knowl. Data Eng.}} \bibinfo{volume}{36}, \bibinfo{number}{11} (\bibinfo{year}{2024}), \bibinfo{pages}{6266--6279}.
\newblock


\bibitem[Kahn et~al\mbox{.}(2021)]%
        {DBLP:journals/ral/KahnAL21}
\bibfield{author}{\bibinfo{person}{Gregory Kahn}, \bibinfo{person}{Pieter Abbeel}, {and} \bibinfo{person}{Sergey Levine}.} \bibinfo{year}{2021}\natexlab{}.
\newblock \showarticletitle{{BADGR:} An Autonomous Self-Supervised Learning-Based Navigation System}.
\newblock \bibinfo{journal}{\emph{{IEEE} Robotics Autom. Lett.}} \bibinfo{volume}{6}, \bibinfo{number}{2} (\bibinfo{year}{2021}), \bibinfo{pages}{1312--1319}.
\newblock
\href{https://doi.org/10.1109/LRA.2021.3057023}{doi:\nolinkurl{10.1109/LRA.2021.3057023}}


\bibitem[Ling and Jacobs(2007)]%
        {DBLP:journals/pami/LingJ07}
\bibfield{author}{\bibinfo{person}{Haibin Ling} {and} \bibinfo{person}{David~W. Jacobs}.} \bibinfo{year}{2007}\natexlab{}.
\newblock \showarticletitle{Shape Classification Using the Inner-Distance}.
\newblock \bibinfo{journal}{\emph{{IEEE} Trans. Pattern Anal. Mach. Intell.}} \bibinfo{volume}{29}, \bibinfo{number}{2} (\bibinfo{year}{2007}), \bibinfo{pages}{286--299}.
\newblock


\bibitem[Liu et~al\mbox{.}(2024)]%
        {DBLP:conf/cvpr/LiuCJQ0P24}
\bibfield{author}{\bibinfo{person}{Yu Liu}, \bibinfo{person}{Yaqi Cai}, \bibinfo{person}{Qi Jia}, \bibinfo{person}{Binglin Qiu}, \bibinfo{person}{Weimin Wang}, {and} \bibinfo{person}{Nan Pu}.} \bibinfo{year}{2024}\natexlab{}.
\newblock \showarticletitle{Novel Class Discovery for Ultra-Fine-Grained Visual Categorization}. In \bibinfo{booktitle}{\emph{{IEEE/CVF} Conference on Computer Vision and Pattern Recognition, {CVPR} 2024, Seattle, WA, USA, June 16-22, 2024}}. \bibinfo{publisher}{{IEEE}}, \bibinfo{pages}{17679--17688}.
\newblock
\href{https://doi.org/10.1109/CVPR52733.2024.01674}{doi:\nolinkurl{10.1109/CVPR52733.2024.01674}}


\bibitem[Liu et~al\mbox{.}(2023)]%
        {DBLP:journals/tkde/LiuJPZZXY23}
\bibfield{author}{\bibinfo{person}{Yixin Liu}, \bibinfo{person}{Ming Jin}, \bibinfo{person}{Shirui Pan}, \bibinfo{person}{Chuan Zhou}, \bibinfo{person}{Yu Zheng}, \bibinfo{person}{Feng Xia}, {and} \bibinfo{person}{Philip~S. Yu}.} \bibinfo{year}{2023}\natexlab{}.
\newblock \showarticletitle{Graph Self-Supervised Learning: {A} Survey}.
\newblock \bibinfo{journal}{\emph{{IEEE} Trans. Knowl. Data Eng.}} \bibinfo{volume}{35}, \bibinfo{number}{6} (\bibinfo{year}{2023}), \bibinfo{pages}{5879--5900}.
\newblock
\href{https://doi.org/10.1109/TKDE.2022.3172903}{doi:\nolinkurl{10.1109/TKDE.2022.3172903}}


\bibitem[Liu et~al\mbox{.}(2021)]%
        {DBLP:conf/iccv/LiuL00W0LG21}
\bibfield{author}{\bibinfo{person}{Ze Liu}, \bibinfo{person}{Yutong Lin}, \bibinfo{person}{Yue Cao}, \bibinfo{person}{Han Hu}, \bibinfo{person}{Yixuan Wei}, \bibinfo{person}{Zheng Zhang}, \bibinfo{person}{Stephen Lin}, {and} \bibinfo{person}{Baining Guo}.} \bibinfo{year}{2021}\natexlab{}.
\newblock \showarticletitle{Swin Transformer: Hierarchical Vision Transformer using Shifted Windows}. In \bibinfo{booktitle}{\emph{2021 {IEEE/CVF} International Conference on Computer Vision, {ICCV} 2021, Montreal, QC, Canada, October 10-17, 2021}}. \bibinfo{publisher}{{IEEE}}, \bibinfo{pages}{9992--10002}.
\newblock
\href{https://doi.org/10.1109/ICCV48922.2021.00986}{doi:\nolinkurl{10.1109/ICCV48922.2021.00986}}


\bibitem[Misra and van~der Maaten(2020)]%
        {DBLP:conf/cvpr/MisraM20}
\bibfield{author}{\bibinfo{person}{Ishan Misra} {and} \bibinfo{person}{Laurens van~der Maaten}.} \bibinfo{year}{2020}\natexlab{}.
\newblock \showarticletitle{Self-Supervised Learning of Pretext-Invariant Representations}. In \bibinfo{booktitle}{\emph{2020 {IEEE/CVF} Conference on Computer Vision and Pattern Recognition, {CVPR} 2020, Seattle, WA, USA, June 13-19, 2020}}. \bibinfo{publisher}{Computer Vision Foundation / {IEEE}}, \bibinfo{pages}{6706--6716}.
\newblock
\href{https://doi.org/10.1109/CVPR42600.2020.00674}{doi:\nolinkurl{10.1109/CVPR42600.2020.00674}}


\bibitem[Ni et~al\mbox{.}(2023)]%
        {DBLP:conf/iccv/NiWYJCH23}
\bibfield{author}{\bibinfo{person}{Zanlin Ni}, \bibinfo{person}{Yulin Wang}, \bibinfo{person}{Jiangwei Yu}, \bibinfo{person}{Haojun Jiang}, \bibinfo{person}{Yue Cao}, {and} \bibinfo{person}{Gao Huang}.} \bibinfo{year}{2023}\natexlab{}.
\newblock \showarticletitle{Deep Incubation: Training Large Models by Divide-and-Conquering}. In \bibinfo{booktitle}{\emph{{IEEE/CVF} International Conference on Computer Vision, {ICCV} 2023, Paris, France, October 1-6, 2023}}. \bibinfo{publisher}{{IEEE}}, \bibinfo{pages}{17289--17299}.
\newblock
\href{https://doi.org/10.1109/ICCV51070.2023.01590}{doi:\nolinkurl{10.1109/ICCV51070.2023.01590}}


\bibitem[Pan et~al\mbox{.}(2023)]%
        {DBLP:conf/wacv/PanYZG23}
\bibfield{author}{\bibinfo{person}{Zicheng Pan}, \bibinfo{person}{Xiaohan Yu}, \bibinfo{person}{Miaohua Zhang}, {and} \bibinfo{person}{Yongsheng Gao}.} \bibinfo{year}{2023}\natexlab{}.
\newblock \showarticletitle{SSFE-Net: Self-Supervised Feature Enhancement for Ultra-Fine-Grained Few-Shot Class Incremental Learning}. In \bibinfo{booktitle}{\emph{{IEEE/CVF} Winter Conference on Applications of Computer Vision, {WACV} 2023, Waikoloa, HI, USA, January 2-7, 2023}}. \bibinfo{publisher}{{IEEE}}, \bibinfo{pages}{6264--6273}.
\newblock
\href{https://doi.org/10.1109/WACV56688.2023.00621}{doi:\nolinkurl{10.1109/WACV56688.2023.00621}}


\bibitem[Patacchiola and Storkey(2020)]%
        {DBLP:conf/nips/PatacchiolaS20}
\bibfield{author}{\bibinfo{person}{Massimiliano Patacchiola} {and} \bibinfo{person}{Amos~J. Storkey}.} \bibinfo{year}{2020}\natexlab{}.
\newblock \showarticletitle{Self-Supervised Relational Reasoning for Representation Learning}. In \bibinfo{booktitle}{\emph{Advances in Neural Information Processing Systems 33: Annual Conference on Neural Information Processing Systems 2020, NeurIPS 2020, December 6-12, 2020, virtual}}, \bibfield{editor}{\bibinfo{person}{Hugo Larochelle}, \bibinfo{person}{Marc'Aurelio Ranzato}, \bibinfo{person}{Raia Hadsell}, \bibinfo{person}{Maria{-}Florina Balcan}, {and} \bibinfo{person}{Hsuan{-}Tien Lin}} (Eds.).
\newblock


\bibitem[Peng et~al\mbox{.}(2018)]%
        {DBLP:journals/tip/PengHZ18}
\bibfield{author}{\bibinfo{person}{Yuxin Peng}, \bibinfo{person}{Xiangteng He}, {and} \bibinfo{person}{Junjie Zhao}.} \bibinfo{year}{2018}\natexlab{}.
\newblock \showarticletitle{Object-Part Attention Model for Fine-Grained Image Classification}.
\newblock \bibinfo{journal}{\emph{{IEEE} Trans. Image Process.}} \bibinfo{volume}{27}, \bibinfo{number}{3} (\bibinfo{year}{2018}), \bibinfo{pages}{1487--1500}.
\newblock


\bibitem[Qaim(2020)]%
        {qaim2020role}
\bibfield{author}{\bibinfo{person}{Matin Qaim}.} \bibinfo{year}{2020}\natexlab{}.
\newblock \showarticletitle{Role of new plant breeding technologies for food security and sustainable agricultural development}.
\newblock \bibinfo{journal}{\emph{Applied Economic Perspectives and Policy}} \bibinfo{volume}{42}, \bibinfo{number}{2} (\bibinfo{year}{2020}), \bibinfo{pages}{129--150}.
\newblock


\bibitem[Radford et~al\mbox{.}(2021)]%
        {DBLP:conf/icml/RadfordKHRGASAM21}
\bibfield{author}{\bibinfo{person}{Alec Radford}, \bibinfo{person}{Jong~Wook Kim}, \bibinfo{person}{Chris Hallacy}, \bibinfo{person}{Aditya Ramesh}, \bibinfo{person}{Gabriel Goh}, \bibinfo{person}{Sandhini Agarwal}, \bibinfo{person}{Girish Sastry}, \bibinfo{person}{Amanda Askell}, \bibinfo{person}{Pamela Mishkin}, \bibinfo{person}{Jack Clark}, \bibinfo{person}{Gretchen Krueger}, {and} \bibinfo{person}{Ilya Sutskever}.} \bibinfo{year}{2021}\natexlab{}.
\newblock \showarticletitle{Learning Transferable Visual Models From Natural Language Supervision}. In \bibinfo{booktitle}{\emph{Proceedings of the 38th International Conference on Machine Learning, {ICML} 2021, 18-24 July 2021, Virtual Event}} \emph{(\bibinfo{series}{Proceedings of Machine Learning Research}, Vol.~\bibinfo{volume}{139})}, \bibfield{editor}{\bibinfo{person}{Marina Meila} {and} \bibinfo{person}{Tong Zhang}} (Eds.). \bibinfo{publisher}{{PMLR}}, \bibinfo{pages}{8748--8763}.
\newblock


\bibitem[Sanakoyeu et~al\mbox{.}(2022)]%
        {DBLP:journals/pami/SanakoyeuMTO22}
\bibfield{author}{\bibinfo{person}{Artsiom Sanakoyeu}, \bibinfo{person}{Pingchuan Ma}, \bibinfo{person}{Vadim Tschernezki}, {and} \bibinfo{person}{Bj{\"{o}}rn Ommer}.} \bibinfo{year}{2022}\natexlab{}.
\newblock \showarticletitle{Improving Deep Metric Learning by Divide and Conquer}.
\newblock \bibinfo{journal}{\emph{{IEEE} Trans. Pattern Anal. Mach. Intell.}} \bibinfo{volume}{44}, \bibinfo{number}{11} (\bibinfo{year}{2022}), \bibinfo{pages}{8306--8320}.
\newblock
\href{https://doi.org/10.1109/TPAMI.2021.3113270}{doi:\nolinkurl{10.1109/TPAMI.2021.3113270}}


\bibitem[Sautier et~al\mbox{.}(2022)]%
        {DBLP:conf/cvpr/SautierPGBBM22}
\bibfield{author}{\bibinfo{person}{Corentin Sautier}, \bibinfo{person}{Gilles Puy}, \bibinfo{person}{Spyros Gidaris}, \bibinfo{person}{Alexandre Boulch}, \bibinfo{person}{Andrei Bursuc}, {and} \bibinfo{person}{Renaud Marlet}.} \bibinfo{year}{2022}\natexlab{}.
\newblock \showarticletitle{Image-to-Lidar Self-Supervised Distillation for Autonomous Driving Data}. In \bibinfo{booktitle}{\emph{{IEEE/CVF} Conference on Computer Vision and Pattern Recognition, {CVPR} 2022, New Orleans, LA, USA, June 18-24, 2022}}. \bibinfo{publisher}{{IEEE}}, \bibinfo{pages}{9881--9891}.
\newblock
\href{https://doi.org/10.1109/CVPR52688.2022.00966}{doi:\nolinkurl{10.1109/CVPR52688.2022.00966}}


\bibitem[Song et~al\mbox{.}(2024)]%
        {DBLP:journals/tip/SongHZCH24}
\bibfield{author}{\bibinfo{person}{Qi Song}, \bibinfo{person}{Qingyong Hu}, \bibinfo{person}{Chi Zhang}, \bibinfo{person}{Yongquan Chen}, {and} \bibinfo{person}{Rui Huang}.} \bibinfo{year}{2024}\natexlab{}.
\newblock \showarticletitle{Divide and Conquer: Improving Multi-Camera 3D Perception With 2D Semantic-Depth Priors and Input-Dependent Queries}.
\newblock \bibinfo{journal}{\emph{{IEEE} Trans. Image Process.}}  \bibinfo{volume}{33} (\bibinfo{year}{2024}), \bibinfo{pages}{897--909}.
\newblock
\href{https://doi.org/10.1109/TIP.2024.3352808}{doi:\nolinkurl{10.1109/TIP.2024.3352808}}


\bibitem[Sun et~al\mbox{.}(2024)]%
        {DBLP:journals/tip/SunHXP24}
\bibfield{author}{\bibinfo{person}{Hongbo Sun}, \bibinfo{person}{Xiangteng He}, \bibinfo{person}{Jinglin Xu}, {and} \bibinfo{person}{Yuxin Peng}.} \bibinfo{year}{2024}\natexlab{}.
\newblock \showarticletitle{{SIM-OFE:} Structure Information Mining and Object-Aware Feature Enhancement for Fine-Grained Visual Categorization}.
\newblock \bibinfo{journal}{\emph{{IEEE} Trans. Image Process.}}  \bibinfo{volume}{33} (\bibinfo{year}{2024}), \bibinfo{pages}{5312--5326}.
\newblock
\href{https://doi.org/10.1109/TIP.2024.3459788}{doi:\nolinkurl{10.1109/TIP.2024.3459788}}


\bibitem[Taleb et~al\mbox{.}(2020)]%
        {DBLP:conf/nips/TalebLDSGBL20}
\bibfield{author}{\bibinfo{person}{Aiham Taleb}, \bibinfo{person}{Winfried Loetzsch}, \bibinfo{person}{Noel Danz}, \bibinfo{person}{Julius Severin}, \bibinfo{person}{Thomas G{\"{a}}rtner}, \bibinfo{person}{Benjamin Bergner}, {and} \bibinfo{person}{Christoph Lippert}.} \bibinfo{year}{2020}\natexlab{}.
\newblock \showarticletitle{3D Self-Supervised Methods for Medical Imaging}. In \bibinfo{booktitle}{\emph{Advances in Neural Information Processing Systems 33: Annual Conference on Neural Information Processing Systems 2020, NeurIPS 2020, December 6-12, 2020, virtual}}, \bibfield{editor}{\bibinfo{person}{Hugo Larochelle}, \bibinfo{person}{Marc'Aurelio Ranzato}, \bibinfo{person}{Raia Hadsell}, \bibinfo{person}{Maria{-}Florina Balcan}, {and} \bibinfo{person}{Hsuan{-}Tien Lin}} (Eds.).
\newblock


\bibitem[Thudi et~al\mbox{.}(2021)]%
        {thudi2021genomic}
\bibfield{author}{\bibinfo{person}{Mahendar Thudi}, \bibinfo{person}{Ramesh Palakurthi}, \bibinfo{person}{James~C Schnable}, \bibinfo{person}{Annapurna Chitikineni}, \bibinfo{person}{Susanne Dreisigacker}, \bibinfo{person}{Emma Mace}, \bibinfo{person}{Rakesh~K Srivastava}, \bibinfo{person}{C~Tara Satyavathi}, \bibinfo{person}{Damaris Odeny}, \bibinfo{person}{Vijay~K Tiwari}, {et~al\mbox{.}}} \bibinfo{year}{2021}\natexlab{}.
\newblock \showarticletitle{Genomic resources in plant breeding for sustainable agriculture}.
\newblock \bibinfo{journal}{\emph{Journal of Plant Physiology}}  \bibinfo{volume}{257} (\bibinfo{year}{2021}), \bibinfo{pages}{153351}.
\newblock


\bibitem[Tian et~al\mbox{.}(2024)]%
        {DBLP:conf/aaai/TianCLYZ24}
\bibfield{author}{\bibinfo{person}{Yanling Tian}, \bibinfo{person}{Di Chen}, \bibinfo{person}{Yunan Liu}, \bibinfo{person}{Jian Yang}, {and} \bibinfo{person}{Shanshan Zhang}.} \bibinfo{year}{2024}\natexlab{}.
\newblock \showarticletitle{Divide and Conquer: Hybrid Pre-training for Person Search}. In \bibinfo{booktitle}{\emph{Thirty-Eighth {AAAI} Conference on Artificial Intelligence, {AAAI} 2024, Thirty-Sixth Conference on Innovative Applications of Artificial Intelligence, {IAAI} 2024, Fourteenth Symposium on Educational Advances in Artificial Intelligence, {EAAI} 2014, February 20-27, 2024, Vancouver, Canada}}, \bibfield{editor}{\bibinfo{person}{Michael~J. Wooldridge}, \bibinfo{person}{Jennifer~G. Dy}, {and} \bibinfo{person}{Sriraam Natarajan}} (Eds.). \bibinfo{publisher}{{AAAI} Press}, \bibinfo{pages}{5224--5232}.
\newblock
\href{https://doi.org/10.1609/AAAI.V38I6.28329}{doi:\nolinkurl{10.1609/AAAI.V38I6.28329}}


\bibitem[Touvron et~al\mbox{.}(2021)]%
        {DBLP:conf/icml/TouvronCDMSJ21}
\bibfield{author}{\bibinfo{person}{Hugo Touvron}, \bibinfo{person}{Matthieu Cord}, \bibinfo{person}{Matthijs Douze}, \bibinfo{person}{Francisco Massa}, \bibinfo{person}{Alexandre Sablayrolles}, {and} \bibinfo{person}{Herv{\'{e}} J{\'{e}}gou}.} \bibinfo{year}{2021}\natexlab{}.
\newblock \showarticletitle{Training data-efficient image transformers {\&} distillation through attention}. In \bibinfo{booktitle}{\emph{Proceedings of the 38th International Conference on Machine Learning, {ICML} 2021, 18-24 July 2021, Virtual Event}} \emph{(\bibinfo{series}{Proceedings of Machine Learning Research}, Vol.~\bibinfo{volume}{139})}, \bibfield{editor}{\bibinfo{person}{Marina Meila} {and} \bibinfo{person}{Tong Zhang}} (Eds.). \bibinfo{publisher}{{PMLR}}, \bibinfo{pages}{10347--10357}.
\newblock


\bibitem[Wang and Gao(2014)]%
        {DBLP:journals/tip/WangG14}
\bibfield{author}{\bibinfo{person}{Bin Wang} {and} \bibinfo{person}{Yongsheng Gao}.} \bibinfo{year}{2014}\natexlab{}.
\newblock \showarticletitle{Hierarchical String Cuts: {A} Translation, Rotation, Scale, and Mirror Invariant Descriptor for Fast Shape Retrieval}.
\newblock \bibinfo{journal}{\emph{{IEEE} Trans. Image Process.}} \bibinfo{volume}{23}, \bibinfo{number}{9} (\bibinfo{year}{2014}), \bibinfo{pages}{4101--4111}.
\newblock


\bibitem[Wang et~al\mbox{.}(2024b)]%
        {DBLP:conf/cvpr/Wang00TPZJ24}
\bibfield{author}{\bibinfo{person}{Chengyao Wang}, \bibinfo{person}{Li Jiang}, \bibinfo{person}{Xiaoyang Wu}, \bibinfo{person}{Zhuotao Tian}, \bibinfo{person}{Bohao Peng}, \bibinfo{person}{Hengshuang Zhao}, {and} \bibinfo{person}{Jiaya Jia}.} \bibinfo{year}{2024}\natexlab{b}.
\newblock \showarticletitle{GroupContrast: Semantic-Aware Self-Supervised Representation Learning for 3D Understanding}. In \bibinfo{booktitle}{\emph{{IEEE/CVF} Conference on Computer Vision and Pattern Recognition, {CVPR} 2024, Seattle, WA, USA, June 16-22, 2024}}. \bibinfo{publisher}{{IEEE}}, \bibinfo{pages}{4917--4928}.
\newblock
\href{https://doi.org/10.1109/CVPR52733.2024.00470}{doi:\nolinkurl{10.1109/CVPR52733.2024.00470}}


\bibitem[Wang et~al\mbox{.}(2024a)]%
        {DBLP:conf/cvpr/WangBWFYYZDZWQY24}
\bibfield{author}{\bibinfo{person}{Yanhui Wang}, \bibinfo{person}{Jianmin Bao}, \bibinfo{person}{Wenming Weng}, \bibinfo{person}{Ruoyu Feng}, \bibinfo{person}{Dacheng Yin}, \bibinfo{person}{Tao Yang}, \bibinfo{person}{Jingxu Zhang}, \bibinfo{person}{Qi Dai}, \bibinfo{person}{Zhiyuan Zhao}, \bibinfo{person}{Chunyu Wang}, \bibinfo{person}{Kai Qiu}, \bibinfo{person}{Yuhui Yuan}, \bibinfo{person}{Xiaoyan Sun}, \bibinfo{person}{Chong Luo}, {and} \bibinfo{person}{Baining Guo}.} \bibinfo{year}{2024}\natexlab{a}.
\newblock \showarticletitle{MicroCinema: {A} Divide-and-Conquer Approach for Text-to-Video Generation}. In \bibinfo{booktitle}{\emph{{IEEE/CVF} Conference on Computer Vision and Pattern Recognition, {CVPR} 2024, Seattle, WA, USA, June 16-22, 2024}}. \bibinfo{publisher}{{IEEE}}, \bibinfo{pages}{8414--8424}.
\newblock
\href{https://doi.org/10.1109/CVPR52733.2024.00804}{doi:\nolinkurl{10.1109/CVPR52733.2024.00804}}


\bibitem[Wang et~al\mbox{.}(2020)]%
        {DBLP:conf/cvpr/WangZKSC20}
\bibfield{author}{\bibinfo{person}{Yude Wang}, \bibinfo{person}{Jie Zhang}, \bibinfo{person}{Meina Kan}, \bibinfo{person}{Shiguang Shan}, {and} \bibinfo{person}{Xilin Chen}.} \bibinfo{year}{2020}\natexlab{}.
\newblock \showarticletitle{Self-Supervised Equivariant Attention Mechanism for Weakly Supervised Semantic Segmentation}. In \bibinfo{booktitle}{\emph{2020 {IEEE/CVF} Conference on Computer Vision and Pattern Recognition, {CVPR} 2020, Seattle, WA, USA, June 13-19, 2020}}. \bibinfo{publisher}{Computer Vision Foundation / {IEEE}}, \bibinfo{pages}{12272--12281}.
\newblock
\href{https://doi.org/10.1109/CVPR42600.2020.01229}{doi:\nolinkurl{10.1109/CVPR42600.2020.01229}}


\bibitem[Wang et~al\mbox{.}(2024c)]%
        {DBLP:journals/corr/abs-2401-15688}
\bibfield{author}{\bibinfo{person}{Zhenyu Wang}, \bibinfo{person}{Enze Xie}, \bibinfo{person}{Aoxue Li}, \bibinfo{person}{Zhongdao Wang}, \bibinfo{person}{Xihui Liu}, {and} \bibinfo{person}{Zhenguo Li}.} \bibinfo{year}{2024}\natexlab{c}.
\newblock \showarticletitle{Divide and Conquer: Language Models can Plan and Self-Correct for Compositional Text-to-Image Generation}.
\newblock \bibinfo{journal}{\emph{CoRR}}  \bibinfo{volume}{abs/2401.15688} (\bibinfo{year}{2024}).
\newblock
\showeprint[arXiv]{2401.15688}
\href{https://doi.org/10.48550/ARXIV.2401.15688}{doi:\nolinkurl{10.48550/ARXIV.2401.15688}}


\bibitem[Wen et~al\mbox{.}(2023)]%
        {DBLP:conf/iccv/Wen0CXS23}
\bibfield{author}{\bibinfo{person}{Yunqian Wen}, \bibinfo{person}{Bo Liu}, \bibinfo{person}{Jingyi Cao}, \bibinfo{person}{Rong Xie}, {and} \bibinfo{person}{Li Song}.} \bibinfo{year}{2023}\natexlab{}.
\newblock \showarticletitle{Divide and Conquer: a Two-Step Method for High Quality Face De-identification with Model Explainability}. In \bibinfo{booktitle}{\emph{{IEEE/CVF} International Conference on Computer Vision, {ICCV} 2023, Paris, France, October 1-6, 2023}}. \bibinfo{publisher}{{IEEE}}, \bibinfo{pages}{5125--5134}.
\newblock
\href{https://doi.org/10.1109/ICCV51070.2023.00475}{doi:\nolinkurl{10.1109/ICCV51070.2023.00475}}


\bibitem[Wu et~al\mbox{.}(2023)]%
        {DBLP:journals/tcsv/WuSXZY23}
\bibfield{author}{\bibinfo{person}{Zhiliang Wu}, \bibinfo{person}{Changchang Sun}, \bibinfo{person}{Hanyu Xuan}, \bibinfo{person}{Kang Zhang}, {and} \bibinfo{person}{Yan Yan}.} \bibinfo{year}{2023}\natexlab{}.
\newblock \showarticletitle{Divide-and-Conquer Completion Network for Video Inpainting}.
\newblock \bibinfo{journal}{\emph{{IEEE} Trans. Circuits Syst. Video Technol.}} \bibinfo{volume}{33}, \bibinfo{number}{6} (\bibinfo{year}{2023}), \bibinfo{pages}{2753--2766}.
\newblock
\href{https://doi.org/10.1109/TCSVT.2022.3225911}{doi:\nolinkurl{10.1109/TCSVT.2022.3225911}}


\bibitem[Xiao et~al\mbox{.}(2023)]%
        {DBLP:conf/icml/0002FZLZ23}
\bibfield{author}{\bibinfo{person}{Jie Xiao}, \bibinfo{person}{Xueyang Fu}, \bibinfo{person}{Man Zhou}, \bibinfo{person}{Hongjian Liu}, {and} \bibinfo{person}{Zheng{-}Jun Zha}.} \bibinfo{year}{2023}\natexlab{}.
\newblock \showarticletitle{Random Shuffle Transformer for Image Restoration}. In \bibinfo{booktitle}{\emph{International Conference on Machine Learning, {ICML} 2023, 23-29 July 2023, Honolulu, Hawaii, {USA}}} \emph{(\bibinfo{series}{Proceedings of Machine Learning Research}, Vol.~\bibinfo{volume}{202})}, \bibfield{editor}{\bibinfo{person}{Andreas Krause}, \bibinfo{person}{Emma Brunskill}, \bibinfo{person}{Kyunghyun Cho}, \bibinfo{person}{Barbara Engelhardt}, \bibinfo{person}{Sivan Sabato}, {and} \bibinfo{person}{Jonathan Scarlett}} (Eds.). \bibinfo{publisher}{{PMLR}}, \bibinfo{pages}{38039--38058}.
\newblock


\bibitem[Xiong et~al\mbox{.}(2023)]%
        {DBLP:journals/ijcv/XiongLLLSC23}
\bibfield{author}{\bibinfo{person}{Haipeng Xiong}, \bibinfo{person}{Hao Lu}, \bibinfo{person}{Chengxin Liu}, \bibinfo{person}{Liang Liu}, \bibinfo{person}{Chunhua Shen}, {and} \bibinfo{person}{Zhiguo Cao}.} \bibinfo{year}{2023}\natexlab{}.
\newblock \showarticletitle{From Open Set to Closed Set: Supervised Spatial Divide-and-Conquer for Object Counting}.
\newblock \bibinfo{journal}{\emph{Int. J. Comput. Vis.}} \bibinfo{volume}{131}, \bibinfo{number}{7} (\bibinfo{year}{2023}), \bibinfo{pages}{1722--1740}.
\newblock
\href{https://doi.org/10.1007/S11263-023-01782-1}{doi:\nolinkurl{10.1007/S11263-023-01782-1}}


\bibitem[Yang et~al\mbox{.}(2022a)]%
        {DBLP:conf/cvpr/YangDTXCCZCH22}
\bibfield{author}{\bibinfo{person}{Jinyu Yang}, \bibinfo{person}{Jiali Duan}, \bibinfo{person}{Son Tran}, \bibinfo{person}{Yi Xu}, \bibinfo{person}{Sampath Chanda}, \bibinfo{person}{Liqun Chen}, \bibinfo{person}{Belinda Zeng}, \bibinfo{person}{Trishul Chilimbi}, {and} \bibinfo{person}{Junzhou Huang}.} \bibinfo{year}{2022}\natexlab{a}.
\newblock \showarticletitle{Vision-Language Pre-Training with Triple Contrastive Learning}. In \bibinfo{booktitle}{\emph{{IEEE/CVF} Conference on Computer Vision and Pattern Recognition, {CVPR} 2022, New Orleans, LA, USA, June 18-24, 2022}}. \bibinfo{publisher}{{IEEE}}, \bibinfo{pages}{15650--15659}.
\newblock
\href{https://doi.org/10.1109/CVPR52688.2022.01522}{doi:\nolinkurl{10.1109/CVPR52688.2022.01522}}


\bibitem[Yang et~al\mbox{.}(2022b)]%
        {DBLP:conf/cvpr/YangZYWD22}
\bibfield{author}{\bibinfo{person}{Muli Yang}, \bibinfo{person}{Yuehua Zhu}, \bibinfo{person}{Jiaping Yu}, \bibinfo{person}{Aming Wu}, {and} \bibinfo{person}{Cheng Deng}.} \bibinfo{year}{2022}\natexlab{b}.
\newblock \showarticletitle{Divide and Conquer: Compositional Experts for Generalized Novel Class Discovery}. In \bibinfo{booktitle}{\emph{{IEEE/CVF} Conference on Computer Vision and Pattern Recognition, {CVPR} 2022, New Orleans, LA, USA, June 18-24, 2022}}. \bibinfo{publisher}{{IEEE}}, \bibinfo{pages}{14248--14257}.
\newblock
\href{https://doi.org/10.1109/CVPR52688.2022.01387}{doi:\nolinkurl{10.1109/CVPR52688.2022.01387}}


\bibitem[Yang et~al\mbox{.}(2023)]%
        {DBLP:conf/cvpr/YangLZW023}
\bibfield{author}{\bibinfo{person}{Zhengwei Yang}, \bibinfo{person}{Meng Lin}, \bibinfo{person}{Xian Zhong}, \bibinfo{person}{Yu Wu}, {and} \bibinfo{person}{Zheng Wang}.} \bibinfo{year}{2023}\natexlab{}.
\newblock \showarticletitle{Good is Bad: Causality Inspired Cloth-debiasing for Cloth-changing Person Re-identification}. In \bibinfo{booktitle}{\emph{{IEEE/CVF} Conference on Computer Vision and Pattern Recognition, {CVPR} 2023, Vancouver, BC, Canada, June 17-24, 2023}}. \bibinfo{publisher}{{IEEE}}, \bibinfo{pages}{1472--1481}.
\newblock
\href{https://doi.org/10.1109/CVPR52729.2023.00148}{doi:\nolinkurl{10.1109/CVPR52729.2023.00148}}


\bibitem[Yao et~al\mbox{.}(2022)]%
        {DBLP:conf/iclr/YaoHHLNXLLJX22}
\bibfield{author}{\bibinfo{person}{Lewei Yao}, \bibinfo{person}{Runhui Huang}, \bibinfo{person}{Lu Hou}, \bibinfo{person}{Guansong Lu}, \bibinfo{person}{Minzhe Niu}, \bibinfo{person}{Hang Xu}, \bibinfo{person}{Xiaodan Liang}, \bibinfo{person}{Zhenguo Li}, \bibinfo{person}{Xin Jiang}, {and} \bibinfo{person}{Chunjing Xu}.} \bibinfo{year}{2022}\natexlab{}.
\newblock \showarticletitle{{FILIP:} Fine-grained Interactive Language-Image Pre-Training}. In \bibinfo{booktitle}{\emph{The Tenth International Conference on Learning Representations, {ICLR} 2022, Virtual Event, April 25-29, 2022}}. \bibinfo{publisher}{OpenReview.net}.
\newblock


\bibitem[Yu et~al\mbox{.}(2023)]%
        {DBLP:conf/ijcai/0001WG23}
\bibfield{author}{\bibinfo{person}{Xiaohan Yu}, \bibinfo{person}{Jun Wang}, {and} \bibinfo{person}{Yongsheng Gao}.} \bibinfo{year}{2023}\natexlab{}.
\newblock \showarticletitle{CLE-ViT: Contrastive Learning Encoded Transformer for Ultra-Fine-Grained Visual Categorization}. In \bibinfo{booktitle}{\emph{Proceedings of the Thirty-Second International Joint Conference on Artificial Intelligence, {IJCAI} 2023, 19th-25th August 2023, Macao, SAR, China}}. \bibinfo{publisher}{ijcai.org}, \bibinfo{pages}{4531--4539}.
\newblock
\href{https://doi.org/10.24963/IJCAI.2023/504}{doi:\nolinkurl{10.24963/IJCAI.2023/504}}


\bibitem[Yu et~al\mbox{.}(2022)]%
        {DBLP:journals/pr/YuZG22}
\bibfield{author}{\bibinfo{person}{Xiaohan Yu}, \bibinfo{person}{Yang Zhao}, {and} \bibinfo{person}{Yongsheng Gao}.} \bibinfo{year}{2022}\natexlab{}.
\newblock \showarticletitle{{SPARE:} Self-supervised part erasing for ultra-fine-grained visual categorization}.
\newblock \bibinfo{journal}{\emph{Pattern Recognit.}}  \bibinfo{volume}{128} (\bibinfo{year}{2022}), \bibinfo{pages}{108691}.
\newblock
\href{https://doi.org/10.1016/J.PATCOG.2022.108691}{doi:\nolinkurl{10.1016/J.PATCOG.2022.108691}}


\bibitem[Yu et~al\mbox{.}(2021a)]%
        {DBLP:journals/pr/YuZGX21}
\bibfield{author}{\bibinfo{person}{Xiaohan Yu}, \bibinfo{person}{Yang Zhao}, \bibinfo{person}{Yongsheng Gao}, {and} \bibinfo{person}{Shengwu Xiong}.} \bibinfo{year}{2021}\natexlab{a}.
\newblock \showarticletitle{MaskCOV: {A} random mask covariance network for ultra-fine-grained visual categorization}.
\newblock \bibinfo{journal}{\emph{Pattern Recognit.}}  \bibinfo{volume}{119} (\bibinfo{year}{2021}), \bibinfo{pages}{108067}.
\newblock
\href{https://doi.org/10.1016/J.PATCOG.2021.108067}{doi:\nolinkurl{10.1016/J.PATCOG.2021.108067}}


\bibitem[Yu et~al\mbox{.}(2020)]%
        {DBLP:conf/aaai/YuZGXY20}
\bibfield{author}{\bibinfo{person}{Xiaohan Yu}, \bibinfo{person}{Yang Zhao}, \bibinfo{person}{Yongsheng Gao}, \bibinfo{person}{Shengwu Xiong}, {and} \bibinfo{person}{Xiaohui Yuan}.} \bibinfo{year}{2020}\natexlab{}.
\newblock \showarticletitle{Patchy Image Structure Classification Using Multi-Orientation Region Transform}. In \bibinfo{booktitle}{\emph{The Thirty-Fourth {AAAI} Conference on Artificial Intelligence, {AAAI} 2020, The Thirty-Second Innovative Applications of Artificial Intelligence Conference, {IAAI} 2020, The Tenth {AAAI} Symposium on Educational Advances in Artificial Intelligence, {EAAI} 2020, New York, NY, USA, February 7-12, 2020}}. \bibinfo{publisher}{{AAAI} Press}, \bibinfo{pages}{12741--12748}.
\newblock
\href{https://doi.org/10.1609/AAAI.V34I07.6968}{doi:\nolinkurl{10.1609/AAAI.V34I07.6968}}


\bibitem[Yu et~al\mbox{.}(2021b)]%
        {DBLP:conf/iccv/YuZ0YX21}
\bibfield{author}{\bibinfo{person}{Xiaohan Yu}, \bibinfo{person}{Yang Zhao}, \bibinfo{person}{Yongsheng Gao}, \bibinfo{person}{Xiaohui Yuan}, {and} \bibinfo{person}{Shengwu Xiong}.} \bibinfo{year}{2021}\natexlab{b}.
\newblock \showarticletitle{Benchmark Platform for Ultra-Fine-Grained Visual Categorization Beyond Human Performance}. In \bibinfo{booktitle}{\emph{2021 {IEEE/CVF} International Conference on Computer Vision, {ICCV} 2021, Montreal, QC, Canada, October 10-17, 2021}}. \bibinfo{publisher}{{IEEE}}, \bibinfo{pages}{10265--10275}.
\newblock
\href{https://doi.org/10.1109/ICCV48922.2021.01012}{doi:\nolinkurl{10.1109/ICCV48922.2021.01012}}


\bibitem[Zhang et~al\mbox{.}(2018)]%
        {DBLP:conf/iclr/ZhangCDL18}
\bibfield{author}{\bibinfo{person}{Hongyi Zhang}, \bibinfo{person}{Moustapha Ciss{\'{e}}}, \bibinfo{person}{Yann~N. Dauphin}, {and} \bibinfo{person}{David Lopez{-}Paz}.} \bibinfo{year}{2018}\natexlab{}.
\newblock \showarticletitle{mixup: Beyond Empirical Risk Minimization}. In \bibinfo{booktitle}{\emph{6th International Conference on Learning Representations, {ICLR} 2018, Vancouver, BC, Canada, April 30 - May 3, 2018, Conference Track Proceedings}}. \bibinfo{publisher}{OpenReview.net}.
\newblock


\bibitem[Zhuang et~al\mbox{.}(2024)]%
        {DBLP:journals/tmm/ZhuangYDQH24}
\bibfield{author}{\bibinfo{person}{Jiamin Zhuang}, \bibinfo{person}{Jing Yu}, \bibinfo{person}{Yang Ding}, \bibinfo{person}{Xiangyan Qu}, {and} \bibinfo{person}{Yue Hu}.} \bibinfo{year}{2024}\natexlab{}.
\newblock \showarticletitle{Towards Fast and Accurate Image-Text Retrieval With Self-Supervised Fine-Grained Alignment}.
\newblock \bibinfo{journal}{\emph{{IEEE} Trans. Multim.}}  \bibinfo{volume}{26} (\bibinfo{year}{2024}), \bibinfo{pages}{1361--1372}.
\newblock
\href{https://doi.org/10.1109/TMM.2023.3280734}{doi:\nolinkurl{10.1109/TMM.2023.3280734}}


\bibitem[Ziegler and Asano(2022)]%
        {DBLP:conf/cvpr/ZieglerA22}
\bibfield{author}{\bibinfo{person}{Adrian Ziegler} {and} \bibinfo{person}{Yuki~M. Asano}.} \bibinfo{year}{2022}\natexlab{}.
\newblock \showarticletitle{Self-Supervised Learning of Object Parts for Semantic Segmentation}. In \bibinfo{booktitle}{\emph{{IEEE/CVF} Conference on Computer Vision and Pattern Recognition, {CVPR} 2022, New Orleans, LA, USA, June 18-24, 2022}}. \bibinfo{publisher}{{IEEE}}, \bibinfo{pages}{14482--14491}.
\newblock
\href{https://doi.org/10.1109/CVPR52688.2022.01410}{doi:\nolinkurl{10.1109/CVPR52688.2022.01410}}


\end{thebibliography}

\end{document}